\pdfoutput=1

\documentclass[11pt]{article}

\usepackage[]{emnlp2022}
\usepackage{times}
\usepackage{latexsym}
\usepackage{amsmath,amssymb}

\usepackage{pifont}
\usepackage[T1]{fontenc}

\usepackage[utf8]{inputenc}

\usepackage{microtype}
\usepackage{inconsolata}

\usepackage{bm}

\usepackage{booktabs}
\usepackage[normalem]{ulem}
\usepackage{graphicx}
\usepackage{bm}
\usepackage{todonotes}
\usepackage{algorithm}
\usepackage{algorithmic}
\usepackage{multirow}
\useunder{\uline}{\ul}{}
\usepackage{hyperref}
\usepackage{verbatim}
\usepackage{lscape}
\usepackage{bbding}
\usepackage{utfsym}
\usepackage{fontawesome}
\usepackage{siunitx}
\newcommand{\hzt}[1]{\textcolor{red}{[hzt: #1]}}

\newcommand{\tenc}[1]{\textcolor{blue}{#1}}
\newcommand{\senc}[1]{\textcolor{red}{#1}}

\title{Composable Text Controls in Latent Space with ODEs}

\makeatletter
\renewcommand{\@fnsymbol}[1]{\ensuremath{\ifcase#1\or \dagger\or \ddagger\or \mathsection\or \mathparagraph\or \|\or **\or \dagger\dagger \or \ddagger\ddagger \else\@ctrerr\fi}}
\makeatother

\author{Guangyi Liu$^{1,3}\thanks{\ \ Work done when Guangyi Liu was a Ph.D. candidate at CUHK-Shenzhen.}$ ,~~
Zeyu Feng$^2$,~~
Yuan Gao$^2$,~~
Zichao Yang$^4$,~~
Xiaodan Liang$^{3,5}$,~~\\
{\bf Junwei Bao$^6$,~~
Xiaodong He$^{6}$,~~
Shuguang Cui$^1$,~~
 Zhen Li$^1$,~~
Zhiting Hu$^{2}$}\\
$^1$FNii, CUHK-Shenzhen,~~ $^2$UC San Diego, ~~$^3$MBZUAI, \\$^4$Carnegie Mellon University, ~~ $^5$DarkMatter AI Research,~~ $^6$JD AI Research\\
{\small \tt guangyi.liu@mbzuai.ac.ae, lizhen@cuhk.edu.cn, zhh019@ucsd.edu}
}

\begin{document}
\maketitle

\begin{abstract}
Real-world text applications often involve \emph{composing} a  wide range of text control operations, such as editing the text \emph{w.r.t.} an attribute, manipulating keywords and structure, and generating new text of desired properties. Prior work typically learns/finetunes a language model (LM) to perform individual or specific subsets of operations. Recent research has studied combining operations in a plug-and-play manner, often with costly search or optimization in the complex sequence space. This paper proposes a new efficient approach for composable text operations in the compact \emph{latent} space of text. The low-dimensionality and differentiability of the text latent vector allow us to develop an efficient sampler based on ordinary differential equations (ODEs) given arbitrary plug-in operators (e.g., attribute classifiers). By connecting pretrained LMs (e.g., GPT2) to the latent space through efficient adaption, we then decode the sampled vectors into desired text sequences. The flexible approach permits diverse control operators (sentiment, tense, formality, keywords, etc.) acquired using any relevant data from different domains. Experiments show that composing those operators within our approach manages to generate or edit high-quality text, substantially improving over previous methods in terms of generation quality and efficiency.\footnote{Code: \url{https://github.com/guangyliu/LatentOps}}
\end{abstract}


\section{Introduction}
\begin{figure}[t]
    \centering
    \includegraphics[width=0.45\textwidth,page=1]{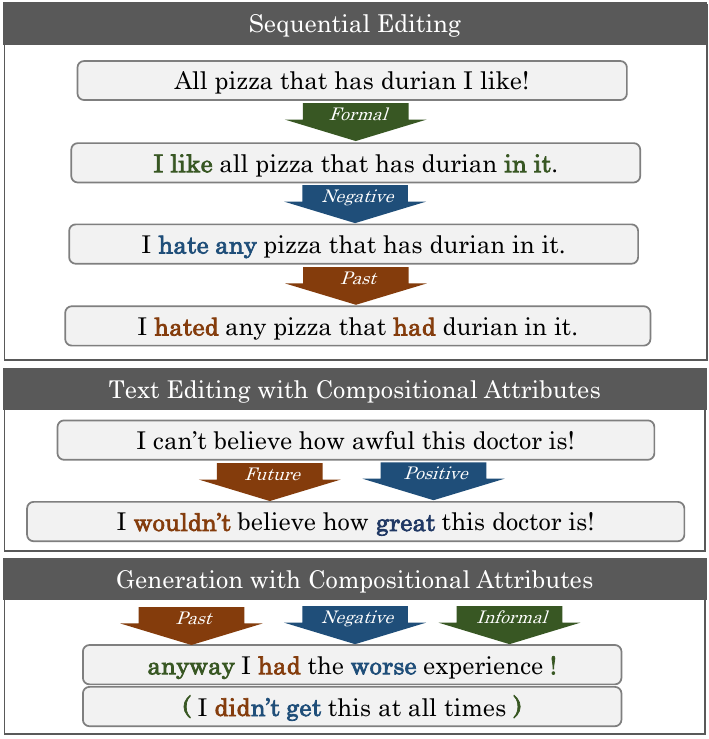}
    \vspace{-8pt}
    \caption{Examples of different composition of text operations, such as editing a text in terms of different attributes sequentially (top) or at the same time (middle), or generating a new text of target properties (bottom). The proposed \textsc{LatentOps} enables a single LM (e.g., an adapted GPT-2) to perform arbitrary text operation composition in the latent space.}
    \label{fig:intro}
    \vspace{-15pt}
\end{figure}

Many text problems involve a diverse set of text control operations, such as editing different attributes (e.g., sentiment, formality) of the text, inserting or changing the keywords, generating new text of diverse properties, and so forth. In particular, different \emph{composition} of those operations are often required in various real-world applications (Figure~\ref{fig:intro}).  

Conventional approaches typically build a conditional model (e.g., by finetuning pretrained language models) for each specific combination of operations~\cite{hu2017toward,keskarCTRL2019,DBLP:journals/corr/abs-1909-08593}, which is unscalable given the combinatorially many possible compositions and the lack of supervised data. Most recent research thus has started to explore plug-and-play solutions. Given a pretrained language model (LM), those approaches plug in arbitrary constraints to guide the production of desired text sequences \cite{Dathathri2020Plug,DBLP:journals/corr/abs-2104-05218, DBLP:conf/nips/KumarMST21, DBLP:conf/emnlp/KrauseGMKJSR21,DBLP:journals/corr/abs-2203-13299,QinCOLD}. 
The approaches, however, typically rely on search or optimization in the complex text \emph{sequence space}. The discrete nature of text makes the search/optimization extremely difficult. Though some recent work introduces continuous approximations to the discrete tokens \citep{qin2020back,QinCOLD,DBLP:conf/nips/KumarMST21}, the high dimensionality and complexity of the sequence space still renders it inefficient to find the accurate high-quality text.

In this paper, we develop \textsc{LatentOps}, a new efficient approach that performs composable control operations in the compact and continuous \emph{latent space} of text. \textsc{LatentOps} permits plugging in arbitrary operators (e.g., attribute classifiers) applied on text latent vectors, to form an energy-based distribution on the low-dimensional latent space. We then develop  
an efficient sampler based on ordinary differential equations (ODEs) \cite{DBLP:conf/iclr/0011SKKEP21,nie2021controllable,Vahdat_LSGM} to draw latent vector samples that bear the desired attributes. 

A key challenge after getting the latent vector is to decode it into the target text sequence. To this end, we connect the latent space to pretrained LM decoders (e.g., GPT-2) by efficiently adapting a small subset of the LM parameters in a variational auto-encoding (VAE) manner~\cite{DBLP:journals/corr/KingmaW13,DBLP:conf/conll/BowmanVVDJB16}. 

Previous attempts of editing text in latent space have often been limited to single attribute and small-scale models, due to the incompatibility of the latent space with the existing transformer-based pretrained LMs \cite{DBLP:conf/nips/WangH019,DBLP:conf/aaai/LiuFZPL20, DBLP:conf/icml/ShenMBJ20,DBLP:conf/acl/DuanXPHL20,DBLP:conf/emnlp/MaiPMSH20}. \textsc{LatentOps} overcomes the difficulties and enables a single large LM to perform arbitrary composable text controls.

We conduct experiments on three challenging settings, including sequential editing of text \emph{w.r.t.} a series of attributes, editing compositional attributes simultaneously, and generating new text given various attributes. Results show that composing operators within our method manages to generate or edit high-quality text, substantially improving over respective baselines in terms of quality and efficiency.

\label{sec:intro}

\section{Background}

\subsection{Energy-based Models and ODE Sampling}
\label{sec:bg_ebms}
Given an arbitrary energy function $E(\bm x)\in \mathbb{R}$, energy-based models (EBMs) define a Boltzmann distribution:
\begin{equation}
\small
\setlength\abovedisplayskip{0.1cm}
\setlength\belowdisplayskip{0.2cm}
    \label{eq:def_ebm}
    p(\bm x) = e^{-E(\bm x)} / Z, 
\end{equation}
where $Z=\sum_{\bm{x} \in \mathcal{X}} e^{-E(\bm x)}$ is the normalization term (the summation is replaced by integration if $\bm x  \in \mathcal{X}$ is a continuous variable). 
EBMs are flexible to incorporate any functions or constraints into the energy function $E(\bm x)$. Recent work has explored text-based EBMs (where $\bm x$ is a text sequence) for controllable text generation \cite{DBLP:conf/nips/HuYSQLDX18,DBLP:conf/iclr/DengBOSR20,DBLP:journals/corr/abs-2106-15078,DBLP:conf/iclr/KhalifaED21,DBLP:journals/corr/abs-2203-13299,QinCOLD}.
Despite the flexibility, sampling from EBMs is rather challenging due to the intractable $Z$. The text-based EBMs
face with even more difficult sampling due to the extremely large and complex (discrete or soft) text space.


Langevin dynamics \cite[LD,][]{DBLP:conf/icml/WellingT11,DBLP:journals/corr/abs-1811-08413} is a gradient-based Markov chain Monte Carlo (MCMC) approach often used for sampling from EBMs \citep{DBLP:conf/nips/DuM19,DBLP:conf/nips/SongE19,DBLP:journals/corr/abs-2004-06030,QinCOLD}. It is considered as a more efficient way compared to other gradient-free alternatives (e.g., Gibbs sampling \citep{bishop2006pattern}). However, due to several critical hyperparameters (e.g., step size, number of steps, noise scale), LD tends to be sensitive and unrobust in practice \citep{nie2021controllable,DBLP:journals/corr/abs-1903-08689,DBLP:conf/iclr/GrathwohlWJD0S20}. 

On the other hand, stochastic/ordinary differential equations (SDEs/ODEs) \cite{anderson1982reverse} offer another sampling technique recently applied in image generation \citep{DBLP:conf/iclr/0011SKKEP21,nie2021controllable}. An SDE characterizes a \emph{diffusion process} that maps real data to random noise in continuous time $t\in[0, T]$. Specifically, let $\bm{x}(t)$ be the value of the process following $\bm{x}(t)\sim p_t(\bm{x})$, indexed by time $t$. At start time $t=0$, $\bm{x}(0)\sim p_0(\bm{x})$ which is the data distribution, and at the end $t=T$, $\bm{x}(T)\sim p_T(\bm{x})$ which is the noise distribution (e.g., standard Gaussian). The \emph{reverse} SDE instead generates a real sample from the noise by working backwards in time (from $t=T$ to $t=0$). More formally, consider a {\it variance-preserving} SDE \citep{DBLP:conf/iclr/0011SKKEP21} whose reverse is written as
\begin{equation}
\small
\setlength\abovedisplayskip{0.2cm}
\setlength\belowdisplayskip{0.2cm}
\label{eq:sde}
    \text{d}\bm x=-\frac{1}{2}\beta(t)[\bm x+2\nabla_{\bm x}\log p_t(\bm x)]\text{d}t + \sqrt{\beta(t)}\text{d}\Bar{\bm w},
\end{equation}
where d$t$ is an infinitesimal negative time step; $\Bar{\bm w}$ is a standard Wiener process when time flows backwards from $T$ to $0$; and the scalar $\beta(t):=\beta_0 + (\beta_T - \beta_0)t$ is a time-variant coefficient linear \emph{w.r.t.} time $t$. Given a noise $\bm{x}(T)\sim p_T(\bm{x})$, solving the above reverse SDE returns a $\bm{x}(0)$ that is a sample from the desired distribution $p_0(\bm{x})$. One could use different numerical solvers to this end.
\cite{burrage2000numerical, higham2001algorithmic,rossler2009second}. 
The SDE sampler sometimes need to combine with an additional corrector to improve the sample quality \citep{DBLP:conf/iclr/0011SKKEP21}.

Further, as shown in \citep{DBLP:conf/iclr/0011SKKEP21,DBLP:journals/entropy/MaoutsaRO20}, each (reverse) SDE has a corresponding ODE, solving which leads to samples following the same distribution. The ODE is written as (see Appendix~\ref{app:derivation_ode} for the derivations):
\begin{equation}
\small
\setlength\abovedisplayskip{0.2cm}
\setlength\belowdisplayskip{0.2cm}
    \label{eq:ode_x}
    \text{d}\bm x=-\frac{1}{2}\beta(t)[\bm x+\nabla_x\log p_t(\bm x)]\text{d}t.
\end{equation}
Solving the ODE with relevant numerical methods~\cite{euler1824institutionum,calvo1990fifth,engstler1997mur8} corresponds to an sampling approach that is more efficient and robust \citep{DBLP:conf/iclr/0011SKKEP21,nie2021controllable}.

In this work, we adapt the ODE sampling for our approach. Crucially, we overcome the text control and sampling difficulties in the aforementioned sequence-space methods, by defining the text control operations in a compact latent space, handled by a latent-space EBMs with the ODE solver for efficient sampling.

\subsection{Latent Text Modeling with Variational Auto-Encoders}
\label{sec:bg_vae}
Variational auto-encoders (VAEs)~\cite{DBLP:journals/corr/KingmaW13,DBLP:conf/icml/RezendeMW14} have been used to model text with a low-dimensional continuous latent space with certain regularities \citep{DBLP:conf/conll/BowmanVVDJB16,hu2017toward}. An VAE connects the text sequence space $\mathcal{X}$ and the latent space $\mathcal{Z}\subset\mathbb{R}^d$ with an encoder $q(\bm z|\bm x)$ that maps text $\bm{x}$ into latent vector $\bm{z}$, and a decoder $p(\bm x|\bm z)$ that maps a $\bm{z}$ into text. Previous work usually learns text VAEs from scratch, optimizing the encoder and decoder parameters with the following objective:
\begin{equation}
\small
\setlength\abovedisplayskip{0.2cm}
\setlength\belowdisplayskip{0.2cm}
\label{eq:vae_loss}
    \begin{split}
        &\mathcal{L}_{\text{VAE}}(\bm x) = \\
        &-\mathbb{E}_{q(\bm z|\bm x)}[\log p(\bm x|\bm z)]
        + \text{KL}(q(\bm z|\bm x) || p_{\text{prior}}(\bm z)),
    \end{split}
\end{equation}
where $p_{\text{prior}}(\bm{z})$ is a standard Gaussian distribution as the prior, and $\text{KL}(\cdot||\cdot)$ is the Kullback-Leibler divergence that pushes $q_{\text{enc}}$ to be close to the prior. The first term encourages $\bm{z}$ to encode relevant information for reconstructing the observed text $\bm{x}$, while the second term adds regularity so that any $\bm{z} \sim p_{\text{prior}}(\bm{z})$ can be decoded into high-quality text in the text sequence space $\mathcal{X}$.
Recent work \citep{li-etal-2020-optimus,hu2021causal} scales up VAE by initializing the encoder and decoder with pretrained LMs (e.g., BERT~\cite{bert} and GPT-2~\cite{gpt2}, respectively). However, they still require costly finetuning of the whole model on the target corpus. 

In comparison, our work converts a given pretrained LM (e.g., GPT-2) into a latent-space model efficiently by tuning only a small subset of parameters, as detailed more in \S\ref{sec:vae_training}.

\begin{figure*}
    \centering
    \vspace{-15pt}
    \includegraphics[width=0.85\textwidth]{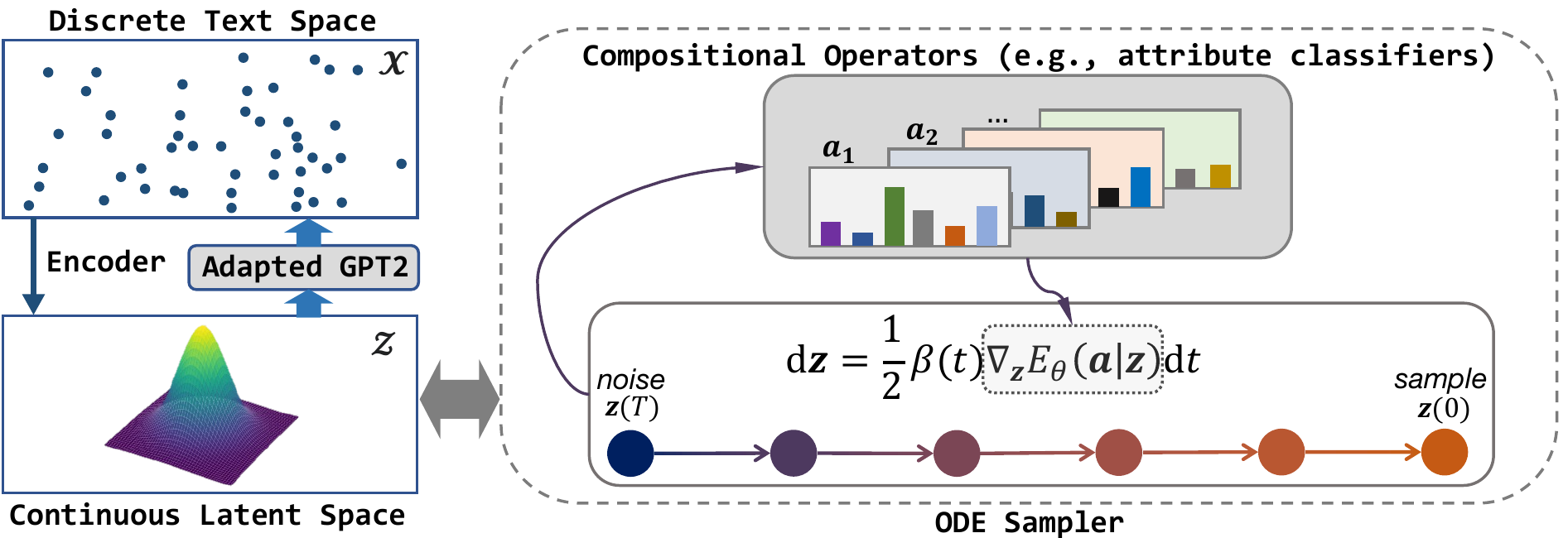}
    \caption{Overview of \textsc{LatentOps}. 
    (Left): We equip pretrained LMs (e.g., GPT-2) with the compact continuous latent space through parameter-efficient adaptation (\S\ref{sec:vae_training}).
    (Right): One could plug in arbitrary operators (e.g., attribute classifiers) to obtain the latent-space EBM (\S\ref{sec:latent_ebms}). We then sample desired latent vectors efficiently by solving the ODE which works backwards through the diffusion process from time $t=T$ to $0$. The resulting sample $\bm{z}(0)$ is fed to the decoder (adapted GPT-2) to generate the desired text sequence.
    }
    \label{fig:model}
    \vspace{-10pt}
\end{figure*}

\section{Composable Text Latent Operations} 
We develop our approach \textsc{LatentOps} that quickly adapts a given pretrained LM (e.g., GPT-2) to enable composable text latent operations. The approach consists of two components, namely a VAE based on the pretrained LM that connects the text space with a compact continuous latent space, and EBMs on the latent space that permits arbitrary attribute composition and efficient sampling. 

More specifically, the VAE decoder $p(\bm{x} | \bm{z})$ offers a way to map any given latent vector $\bm{z}$ into the corresponding text sequence. Therefore, text control (e.g., editing a text or generating a new one) boils down to finding the desired vector $\bm{z}$ that bears the desired attributes and characteristics. To this end, one could plug in any relevant attribute operators (e.g., classifiers), resulting in a latent-space EBM that characterizes the distribution of $\bm{z}$ with the desired attributes. We could then draw the $\bm{z}$ samples of interest, performed efficiently with an ODE solver. Figure~\ref{fig:model} gives an illustration of the approach.

\textsc{LatentOps} thus avoids the difficult optimization or sampling in the complex text sequence space as compared to the previous plug-and-play methods \cite[e.g.,][]{DBLP:journals/corr/abs-2104-05218,Dathathri2020Plug,QinCOLD}. Our approach is also compatible with the powerful pretrained LMs, requiring only minimal adaptation to equip the LMs with a latent space, rather than costly retraining from scratch as in the recent diffusion LM \cite{li2022diffusion}. 

In the following, we first present the latent-space EBM formulation (\S\ref{sec:latent_ebms}) for composable operations, and derive the efficient ODE sampler (\S\ref{sec:ode_sampler}); we discuss the parameter-efficient adaptation of pretrained LMs for the latent space (\S\ref{sec:vae_training});  we then discuss the implementation details (\S\ref{sec:implement}).



\subsection{Composable Latent-Space EBMs}
\label{sec:latent_ebms}

We aim to formulate the latent-space EBMs such that one can easily plug in arbitrary attribute operators to define the latent distribution of interest. Besides, as we want to obtain fluent text with the VAE decoder $p(\bm{x}|\bm{z})$ described in \S\ref{sec:vae_training}, the latent distribution over $\bm{z}$ should match the structure of the VAE latent space. 

Formally, let $\bm a=\{a_1, a_2, ...\}$ be a vector of desired attribute values, where each $a_i \in \mathbb{R}$ (e.g., positive sentiment, or informal writing style). Note that $\bm{a}$ does not have a prefixed length as one can plug in any number of attributes to control on the fly. In general, to assess if a vector $\bm{z}$ bears the desired attribute $a_i$, we could use any function $f_i$ that takes in $\bm{z}$ and $a_i$, and outputs a score measuring how well $a_i$ is carried in $\bm{z}$. 
For a categorical attribute (e.g., sentiment, either positive or negative), one of the common choices is to use a trained attribute classifier, where $f_i(\bm{z})$ is the output logit vector and $f_i(\bm{z})[a_i] \in \mathbb{R}$ is the logit of the particular class $a_i$ of interest. For clarity of presentation, we focus on categorical attributes and classifiers in the rest of the paper, and assume the attributes are independent with each others. 

We are now ready to formulate the latent-space EBMs by plugging in the attribute classifiers. Specifically, we define the joint distribution:
\begin{equation}
\small
\setlength\abovedisplayskip{0.2cm}
\setlength\belowdisplayskip{0.2cm}
\label{eq:joint_dist}
    p(\bm z,\bm a) := p_{\text{prior}}(\bm z) p(\bm a | \bm z)  = p_{\text{prior}}(\bm z) \cdot e^{- E(\bm{a}|\bm z)} / Z,
\end{equation}
where $p_{\text{prior}}(\bm{z})$ is the Gaussian prior distribution of VAE (\S\ref{sec:bg_vae}), and $p(\bm{a}|\bm{z})$ is formulated with energy function $E(\bm{a}|\bm z)$ to encode the different target attributes. Such a decomposition of $p(\bm z, \bm a)$ results in two key desirable properties: (1)
The marginal distribution over $\bm{z}$ equals the VAE prior, i.e., $\sum_{\bm{a}} p(\bm{z}, \bm{a}) = p_{\text{prior}}(\bm{z})$. This facilitates the VAE decoder to generate fluent text;
(2) the energy function in $p(\bm a | \bm z)$ enables the combination of arbitrary attributes, with $E(\bm{a}|\bm z) = \sum_{i} \lambda_i E_i(a_i|\bm z)$. Each $\lambda_i \in \mathbb{R}$ is the balance weight, and $E_i$ is the defined as the negative log probability (i.e., the normalized logit) of $a_i$ to make sure the different attribute classifiers have outputs at the same scale for combination:
\begin{equation}
\small
\setlength\abovedisplayskip{0.2cm}
\setlength\belowdisplayskip{0.01cm}
\label{eq:cond_e_discrete}
    E_i(a_i|\bm z) = -{f_{i}(\bm z)}[a_i] + \log\sum\nolimits_{a'_i}\exp(f_{i}(\bm z)[a'_i]).
\end{equation}

\subsection{Efficient Sampling with ODEs}
\label{sec:ode_sampler}
Once we have the desired distribution $p(\bm z, \bm a)$ over the latent space and attributes, we would like to draw samples $\bm{z}$ given the target attribute values $\bm{a}$. The samples can then be fed to the VAE decoder (\S\ref{sec:vae_training}) to obtain the desired text. As discussed in \S\ref{sec:bg_ebms} and also shown in our ablation study in \S\ref{app:compare_sde}, sampling with ODEs has the benefits of robustness compared to Langevin dynamics that is sensitive to hyperparameters, and efficiency compared to SDEs that require additional correction. 

We now derive the ODE sampling in the latent space. Specifically, we adapt the ODE from Eq.\eqref{eq:ode_x} into our latent-space setting, which gives:
\begin{equation}
\small
\setlength\abovedisplayskip{0.2cm}
\setlength\belowdisplayskip{0.2cm}
    \label{eq:ode_z}
    \begin{split}
        &\text{d}\bm z=-\frac{1}{2}\beta(t)[\bm z+\nabla_{\bm z}\log p_t(\bm z,\bm a)]\text{d}t \\
        &=-\frac{1}{2}\beta(t)\left [\bm z +\nabla_{\bm z}\log p_t(\bm a|\bm z)  + \nabla_{\bm z}\log p_t(\bm z)  \right ] \text{d}t.
    \end{split}
\end{equation}
For $p_t(\bm z)$, notice that at $t=0$, $p_0(\bm{z})$ is the VAE prior distribution $p_{\text{prior}}(\bm z)$ as defined in Eq.\eqref{eq:joint_dist}, which is the same as $p_T(\bm{z})$ (i.e., the Gaussian noise distribution after diffusion). This means that in the diffusion process, we always have $p_t(\bm{z})=\mathcal{N}(\bm 0,I)$ that is time-invariant \citep{nie2021controllable}. Similarly, for $p_t(\bm{a} | \bm{z})$, since the input $\bm{z}$ follows the time-invariant distribution and the classifiers $f_i$ are fixed, the $p_t(\bm a|\bm z)$ is also time-invariant. Plugging the definitions of those components, we obtain the simple ODE formulation:
\begin{equation}
\small
\setlength\abovedisplayskip{0.2cm}
\setlength\belowdisplayskip{0.1cm}
\label{eq:final_ode}
\begin{split}
    \text d\bm z&=-\frac{1}{2}\beta(t)[\bm z-\nabla_{\bm z} E (\bm a|\bm z)- \frac{1}{2}\nabla_{\bm z}||\bm z||^2_2]\text dt\\
    & = \frac{1}{2}\beta(t)\sum_{i=1}^n \nabla_{\bm z} E(a_i|\bm z) \text dt.
\end{split}
\end{equation}
We can then easily create latent samples conditioning on the given attribute values, by drawing $\bm{z}(T) \sim \mathcal{N}(\bm 0,I)$ and solving the Eq.\eqref{eq:final_ode} with a differentiable neural ODE solver\footnote{\url{https://github.com/rtqichen/torchdiffeq}}~\cite{chen2018neuralode,chen2021eventfn} to obtain $\bm{z}(0)$. In \S\ref{sec:implement}, we discuss more implementation details with approximated starting point $\bm{z}(T)$ for text editing and better empirical performance.

\subsection{Adapting Pretrained LMs for Latent Space}
\label{sec:vae_training}
To decode the $\bm{z}$ samples into text sequences, we equip pretrained LMs (e.g., GPT-2) with the latent space through parameter-efficient adaptation. More specifically, we adapt the autoregressive LM into a text latent model within the VAE framework (\S\ref{sec:bg_vae}). Differing from the previous VAE work that trains from scratch or finetunes the full parameters of pretrained LMs \citep{li-etal-2020-optimus,hu2021causal,hu2017toward}, we show that it is sufficient to only update a small portion of the LM parameters to connect the LM with the latent space, while keeping the LM capability of generating fluent coherent text. Specifically, we 
augment the autoregressive LM with small MLP layers that pass the latent vector $\bm{z}$ to the LM, and insert an additional transformer layer in between the LM embedding layer and the original first layer. The resulting model then serves as the decoder in the VAE objective (Eq.\ref{eq:vae_loss}), for which we only optimize the MLP layers, the embedding layer, and the inserted transformer layer, while keeping all other parameters frozen. For the encoder, we use a BERT-small model \cite{bert,DBLP:journals/corr/abs-1908-08962} and finetune it in the VAE framework. As discussed later in \S\ref{sec:implement}, the tuned encoder can be used to produce the initial $\bm z$ values in the ODE sampler for text editing.



\subsection{Implementation Details}\label{sec:implement}

We discuss more implementation details of the method. Overall, given an arbitrary text corpus (e.g., a set of text from any domain of interest), we first build the VAE by adapting the pretrained LMs as described in \S\ref{sec:vae_training}. Once the latent space is established, we keep it (including all the VAE components) fixed, and perform compositional text operations in the latent space on the fly.

\paragraph{Acquisition of attribute classifiers}
We can acquire attribute classifiers $f_i(\bm{z})$ on the frozen latent space by training using arbitrary datasets with annotations. Specifically, we encode the input text into the latent space with the VAE encoder, and then train the classifier to predict the attribute label given the latent vector. Each classifier, as is built on the semantic latent space, can be trained efficiently with only a small number of examples (e.g., 200 per class). This allows us to acquire a large diversity of classifiers (e.g., sentiment, formality, different keywords) in our experiments (\S\ref{sec:exp}) using readily-available data from different domains, and flexibly compose them together to perform operations on text in the domain of interest.

\paragraph{Initialization of ODE sampling}

To sample $\bm{z}$ with the ODE solver (\S\ref{sec:ode_sampler}), we need to specify the initial $\bm{z}(T)$. For text editing operations (e.g., transferring sentiment from positive to negative) that start with a given text sequence, we initialize $\bm{z}(T)$ to the latent vector of the given text by the VAE encoder. We show in our experiments that the resulting $\bm{z}(0)$ samples as the solution of the ODEs can preserve the relevant information in the original text while obtaining the desired target attributes.

For generating new text of target attributes, the normal way is to sample $\bm{z}(T)$ from the prior Gaussian distribution $\mathcal{N}(\bm 0, I)$. However, due to the inevitable gap between the prior distribution and the learned VAE posterior on $\mathcal{Z}$, such a Gaussian noise sample does not always lead to coherent text outputs. We thus follow \citep{li-etal-2020-optimus,hu2021causal} to learn a small (single-layer) GAN \citep{goodfellow2014generative} $p_{\text{GAN}}(\bm z)$ that simulates the VAE posterior distribution, using all encoded $\bm z$ of real text as the training data. We then generate the initial $\bm{z}(T)$ from the $p_{\text{GAN}}$.



\paragraph{Sample selection}
The compact latent space learned by VAE allows us to conveniently create multiple semantically-close variants of a sampled $\bm{z}(0)$ and pick the best one in terms of certain task criteria. Specifically, we add random Gaussian noise perturbation (with a small variance) to $\bm{z}(0)$ to get a set of vectors close to $\bm{z}(0)$ in the latent space and select one from the set. We found the sample perturbation and selection is most useful for operations related to the text content. For example, in text editing (\S\ref{sec:tst_exp}), we pick a vector based on the content preservation (e.g., BLEU with the original text) and attribute accuracy. More details are provided in \S\ref{app:sample_selection}.

\section{Experiments}\label{sec:exp}
We conduct extensive experiments of composable text controls to show the flexibility and efficiency of \textsc{LatentOps}, including generating new text of compositional attributes (\S\ref{sec:cg_exp}) and editing existing text in terms of desired attributes sequentially or simultaneously (\S\ref{sec:tst_exp}). 
All code will be released upon acceptance.
\begin{table}[t]
\vspace{-15pt}
\setlength\tabcolsep{3.4pt}
\scriptsize
\centering
\begin{tabular}{llcccccc}
\toprule
\multirow{3}{*}{Attributes} &\multirow{3}{*}{Methods} & \multicolumn{4}{c}{Accuracy$\uparrow$} &  Fluency$\downarrow$ & Diversity$\downarrow$\\\cmidrule(r){3-6}\cmidrule(r){7-7}\cmidrule(r){8-8}
& &S & T & F & G-M &PPL & sBL    \\
\midrule
\multirow{5}{*}{S}
                 & GPT2-FT& 0.98 & -&-&0.98 & 10.6 & 23.8\\\cmidrule{2-8}
                 & PPLM &0.86&-&-&0.86&11.8&31.0 \\
                & FUDGE&0.77&-&-&0.77&\textbf{10.3}&27.2\\
                 & Ours &\textbf{0.99} &-&-&\textbf{0.99} & 30.4 & \textbf{13.0} \\
                 \midrule
\multirow{5}{*}{S+T} 
                 & GPT2-FT   &0.98&0.95&-&0.969&9.0&36.8 \\\cmidrule{2-8}
                 & PPLM&0.81&0.59&-&0.677&15.7&28.7 \\
                & FUDGE&0.67&0.63&-&0.565&\textbf{11.0}&35.9\\
                 & Ours &\textbf{0.98}& \textbf{0.93}& -&\textbf{0.951}& 25.2& \textbf{19.7}\\
                 \midrule
\multirow{5}{*}{S+T+F} 
                 & GPT2-FT  &0.97&0.92&0.87&0.919&10.3&36.8 \\\cmidrule{2-8}
                 & PPLM &0.82&0.57&0.56&0.598&17.5&30.5  \\
                &FUDGE&0.67&0.64&0.62&0.556&\textbf{11.5}&35.9\\
                 & Ours &\textbf{0.97}& \textbf{0.92}& \textbf{0.93}       &\textbf{0.937}   & 25.8            & \textbf{21.1}\\
                 \midrule
\end{tabular}
\caption{Results of generation with compositional attributes. S, T and F stand for sentiment, tense and formality, respectively.  G-M is the geometric mean of all accuracy. 
  For reference, the PPL of test data and human-annotated data is 15.9 and 24.5. Since GPT2-FT is a fully-supervised model for reference, we mark the best result \textbf{bold} except GPT2-FT.
}
\label{tab:multi_cg}
\vspace{-10pt}
\end{table}
\paragraph{Setup}
We evaluate in two domains, including the Yelp review  \cite{DBLP:conf/nips/ShenLBJ17} preprocessed by \citet{DBLP:conf/naacl/LiJHL18} and the Amazon comment corpus \cite{DBLP:conf/www/HeM16}. 
For each domain, we quickly adapt the GPT2-large to equip with a latent space as described in \S\ref{sec:vae_training}. 
The resulting VAE models then serve as the base model, on which we plug in various attribute classifiers for generation and editing. 
We consider the attributes of \emph{sentiment} (positive, negative), \emph{formality} (formal, informal), and \emph{tense} (pase, present, future). 
(We also study other attributes related to diverse \emph{keywords}, which we present in \S\ref{app:generation_keyword}).
The sentiment/tense classifiers are quickly acquired by training on a small subset of Yelp and Amazon instances (200 labels per class), where the sentiment labels were readily available in the corpus and the tense labels are automatically parsed (\S\ref{app:setup}). 
There is no formality information in the Yelp/Amazon corpora, yet the flexibility of \textsc{LatentOps} allows us to acquire the formality classifier using a separate dataset GYAFC \cite{DBLP:conf/naacl/RaoT18}. 
\S\ref{app:setup} gives more details of the setup.

\subsection{Generation with Compositional Attributes}
\label{sec:cg_exp}

We apply \textsc{LatentOps} to generate new text of arbitrary desired attributes on Yelp domain.

\paragraph{Baselines}
We compare with the previous plug-and-play text control approaches {\bf PPLM}~\cite{Dathathri2020Plug} and {\bf FUDGE}~\cite{DBLP:journals/corr/abs-2104-05218}. 
As mentioned earlier, both approaches apply attribute classifiers on the complex sequence space, with an autoregressive LM as a base model. 
We obtain the base model by finetuning GPT2-large on the above domain corpus (e.g., Yelp). We further compare with an expensive supervised method {\bf GPT2-FT} which finetunes a GPT2-large for \emph{each} combination of attributes. 
To get the supervised data (\S\ref{app:generation_baselines}), we automatically annotate the domain corpus for formality and tense labels with a trained classifier and tagger, respectively. 
\paragraph{Metrics}
Attribute accuracy is given by a BERT classifier to evaluate the success rate.
Perplexity (PPL) is calculated by a GPT2 finetuned on the corresponding domain to measure fluency. 
We calculate self-BLEU (sBL) to evaluate the diversity.
For each case, we sample 150 sequencs to evaluate. 

\subsubsection{Experimental Results}
\label{sec:generation_composable}
We list the average results of each combination in Table~\ref{tab:multi_cg}. \textsc{LatentOps} achieves observably higher accuracy and diversity, even compared with the fully-supervised method (i.e., GPT2-FT). For fluency, the perplexity of our \textsc{LatentOps} is within a regular interval (the perplexity of human-annotated data is 24.5). However, the baselines obtain excessive perplexity at the expense of diversity.
\begin{table}[t]
\centering
\scriptsize
\vspace{-15pt}
\begin{tabular}{m{0.45\textwidth}}
\toprule
\textbf{Negative + Future + Formal}\\
\midrule
GPT2-FT: \\
	\quad i will not be back.\\
	\quad would not recommend this location to anyone. \senc{[No Subject]}\\
	\quad would not recommend them for any jewelry or service. \senc{[No Subject]}\\
	\quad if i could give this place zero stars, i would.\\
	\midrule

PPLM:\\
	\quad i \senc{could} not recommend them at all.\\
	\quad i \senc{could not} believe this \senc{was not good}!\\
	\quad this \senc{was a big deal}, because the food \senc{was great}. \\
	\quad i \senc{could} not recommend them.\\
	\midrule

FUDGE:\\
    \quad not a great pizza to get a great pie! \senc{[No Tense]}\\
    \quad however, this place \senc{is pretty good}. \\
    \quad i \senc{have never} seen anything like these. \\
    \quad will definitely return. \senc{[No Subject]}\\
	\midrule

Ours:\\
	\quad i would not believe them to stay .\\
	\quad i will never be back .\\
	\quad i would not recommend her to anyone in the network .\\
	\quad they will not think to contact me for any reason .\\
	\bottomrule
\end{tabular}
\caption{Examples of generation with compositional attributes. 
We mark failed spans in \senc{red}.
}
\label{tab:examples_compositional}
\vspace{-10pt}
\end{table}

Table~\ref{tab:examples_compositional} shows some generated samples. Ours yields fluent sentences that mostly satisfy the controls. Moreover, GPT2-FT performs similar, although it misses the subject in the second and the third examples. PPLM may fail due to the lack of global concern, e.g., the double negation leads to positive sentiment in the second example. Both PPLM and FUDGE could hardly succeed in all the controls simultaneously since it operates on the sequence space of an autoregressive LM, which is arduous to coordinate the controls.
Refer to \S\ref{app:generation_compositional} for more generated examples and analysis.

\subsubsection{Runtime Efficiency}
\label{sec:runtime}
To quantify the computational cost of each method, we evaluate the consumed time for generating 150 examples.
We start timing after the models are loaded and before the generation starts. And we end timing right after 150 sentences are generated. We run five times for each method and average the results as final results, shown in Table~\ref{tab:time_comsumed}. Since we sample in the low-dimensional compact latent space, our method is 6.6$\times$ faster than FUDGE and 578$\times$ faster than PPLM. 
\begin{table}[t]
\small
    \centering
    \vspace{-20pt}
    \begin{tabular}{cccc}
    \toprule
    Methods & PPLM & FUDGE & Ours \\\midrule
    Time (s) & 3182 (578$\times$) & 36.1 (6.6$\times$) & 5.5 (1$\times$)\\\bottomrule
    \end{tabular}
    \caption{Results of generation time of each method.}
    \label{tab:time_comsumed}
    \vspace{-10pt}
\end{table}

\subsection{Text Editing}
\label{sec:tst_exp}
We evaluate our model's text editing ability on both Yelp and Amazon domains, i.e, changing sentences' sentiment, tense and formality attributes sequentially (\S\ref{sec:sequential_edit}) or altogether (\S\ref{sec:text_edit_compositional}). 

\paragraph{Baselines}
Since few previous works can handle the sequential and compositional attributes editing task, we mainly compare with  FUDGE~\cite{DBLP:journals/corr/abs-2104-05218}. 
Moreover, we train three Style Transformer~\cite{DBLP:conf/acl/DaiLQH19} models (for sentiment, tense, and formality, respectively) to sequentially edit the source sentences as a baseline of sequential editing. 
To show the superiority of our \textsc{LatentOps}, we also conduct text editing with single attribute and compare with several recent state-of-the-art methods (\S\ref{app:baseline_text_edit}).
We adopt the same setting (few-shot) as in \S\ref{sec:cg_exp} for FUDGE and our \textsc{LatentOps}.
It is noteworthy that \textsc{LatentOps} is precisely the same model as in \S\ref{sec:cg_exp}, so it does not require further training. 

\paragraph{Metrics}
Besides success rate and fluency mentioned in \S\ref{sec:cg_exp}, we evaluate the ability of content preservation. Since it is a critical measure lying in the field of text editing,
we utilize two metrics: input-BLEU (iBL, BLEU between input and output)
and CTC score~\cite{ctc_score} (bi-directional information alignment between input and output).
For single attribute setting, we also evaluate reference-BLEU (rBL, BLEU between human-annotated ground truth and output) and perform human evaluations (\S\ref{app:example_text_edit_single}). 

\subsubsection{Sequential Editing}
\label{sec:sequential_edit}
In this section, we give the results of sequential editing, whose goal is to edit the given text by changing an attribute each time and keep the main content consistent. We consider the situation that source sentences are with formal manner, positive sentiment and present tense (selected by external classifiers in Yelp), and the goal is to transfer the source sentences to informal manner, negative sentiment and past tense, separately and sequentially. Potential entanglements exist among these attributes, and it is hard to control each attribute independently.

 The automatic evaluation results are listed in Table~\ref{tab:seq_edit_auto}. 
\textsc{LatentOps} performs the best on acquiring desired controls and maintaining others and achieves a balanced trade-off among accuracy, content alignment, and fluency. 
FUDGE fails to introduce the informal manner, while it achieves better formality controls after introducing negative sentiment, showing its deficiency of ability of disentanglement. Furthermore, although FUDGE preserves the most content, it mistakes the core and puts the cart (content) before the horse (accuracy). STrans performs plain overall and cannot guarantee fluency well. 

\begin{table}[t]
    \centering
    \scriptsize
    \vspace{-20pt}
    \setlength\tabcolsep{4.5pt}
    \begin{tabular}{llcccccc}
    \toprule
    \multirow{3}{*}{Attributes}&\multirow{3}{*}{Methods}&\multicolumn{3}{c}{Accuracy}&\multicolumn{2}{c}{Content$\uparrow$}&Fluency$\downarrow$\\ \cmidrule(r){3-5}  \cmidrule(r){6-7}  \cmidrule(r){8-8} 
     &    & F & S & T & iBL & CTC  &PPL\\\midrule
        \multirow{3}{*}{Informal}& FUDGE & 0.04& 0.06& 0.0 &\textbf{99.4}&0.479&\textbf{19.3}  \\
        & STrans & 0.45      & 0.14      & {0.06}  & {65.4}  & 0.470& 36.0 \\
    & Ours   & \textbf{0.85}      & {0.07}      & 0.07  & 64.2  & \textbf{0.482}& {20.2} \\\midrule
    \multirow{3}{*}{+ Negative}&  FUDGE &   0.49    & 0.35     &  0.10 & \textbf{48.6} & 0.451 & {35.0} \\
    &STrans & 0.38      & 0.82      & 0.10   & {42.4} & 0.457 & 39.9 \\
    & Ours   & \textbf{0.75}      & \textbf{0.92}      & {0.07}  & 42.1 & \textbf{0.468}& \textbf{28.7}  \\\midrule
    \multirow{3}{*}{\ \ + Present} & FUDGE &  0.48    & 0.35      &  0.10 & \textbf{49.3} & {0.452}& \textbf{30.7}  \\
    & STrans & 0.36      & 0.81      & 0.50   & {25.6}  & 0.453& 45.4 \\
    & Ours   & \textbf{0.61}      & \textbf{0.83}      & \textbf{0.74}  & 20.7 & \textbf{0.461}&{31.5}\\\bottomrule
    \end{tabular}
    \caption{Automatic evaluations of sequential editing on Yelp review dataset. F, S and T stand for the accuracy of formality (to informal), sentiment (to negative) and tense (to present), respectively.
    }
    \label{tab:seq_edit_auto}
    \vspace{-10pt}
\end{table}
We provide some examples in Table~\ref{tab:seq_edit_part}. The formality control of FUDGE makes no effect. Besides, FUDGE would introduce some irrelevant information, e.g., \textit{garlic pizza} and \textit{thing's}. A similar situation exists in STrans, e.g., \textit{ate} and \textit{korean food}.  
More examples and analysis are in \S\ref{app:example_seq_edit}.

\subsubsection{Text Editing with Compositional Attributes}
\label{sec:text_edit_compositional}
We give the results of text editing with compositional attributes on Yelp, aiming to edit attributes of sentiment and tense of the source sentences. 
The automatic evaluation results are listed in Table~\ref{tab:tst_multiple}. 
\textsc{LatentOps} achieves a higher success rate and content alignment (CTC). 
FUDGE performs better on iBL and worse on CTC. As demonstrated by \citet{ctc_score}, the two-way approach (CTC) is more effective and exhibits a higher correlation than single-directional alignment (e.g., BLEU), which is consistent with our observation: FUDGE prefers to generate long sentences that contain the spans in source (raise iBL), but it will also introduce irrelevant information (lower CTC). We give some examples in \S\ref{app:example_text_edit_compositional} to support the claim. 
\subsection{Ablation Study}
To clarify the advantage of sampling from ODE, we compare different sampling methods, including Stochastic Gradient Langevin Dynamics (SGLD) and Predictor-Corrector sampler with SDE in \S\ref{app:compare_sde}.

\begin{table}[t]
    \centering
    \vspace{-15pt}
    \setlength\tabcolsep{2.pt}
    \scriptsize
    \begin{tabular}{ll}
    \toprule
         Source & the flowers and prices were great . \\
         \midrule
           FUDGE:&\\
         + informal&the flowers and prices were great. \senc{[Formal]}\\
         \quad + negative& \senc{garlic pizza} and prices were \senc{great}.\\
         \quad\quad + present&\senc{garlic pizza} and prices \senc{were great}.\\
          STans:&\\
         + informal&the flowers and prices were great ?\\
         \quad+ negative&the \senc{ate} and prices were terrible ?\\
         \quad\quad+ present&the \senc{ate} and prices are terrible ?\\
          Ours:& \\
         + informal& and the flowers and prices were great ! \\
         \quad+ negative& and the flowers and prices were terrible !\\
         \quad\quad+ present& and the flowers and prices are terrible !\\\midrule
         Source & best korean food on this side of town .\\\midrule
         FUDGE:&\\
         + informal&best korean food on this side of town. \senc{[Formal]}\\
         \quad+ negative&\senc{thing's best} korean food on this side of town.\\
         \quad\quad+ present &\senc{thing's best} korean food on this side of town. \senc{[No Tense]}\\
          STans:&\\
         + informal&best korean food on this side of town \senc{korean food} . \senc{[Formal]}\\
         \quad+ negative&only korean food on this side of town \senc{korean food} .\\
         \quad\quad+ present&only korean food on this side of town \senc{korean food} . \senc{[No Tense]}\\
          Ours:& \\
          + informal&best korean food on this side of town ! \\
         \quad + negative& worst korean food on this side of town ! \\
          \quad\quad+ present& this is worst korean food on this side of town !\\\bottomrule
    \end{tabular}
    \caption{Some examples of sequential editing. We mark failed spans in \senc{red}. }
    \label{tab:seq_edit_part}
    \vspace{-10pt}
\end{table}

\section{Related Work}
Recent works on text generation can be divided into two categories. One generates desirable texts by directly modifying the text sequence space. The other operates on the latent space to obtain a representation that can be decoded into sequence with desired attributes. More detailed discussions can be found in Section \S\ref{app:lace}.
\subsection{Text Control in Sequence Space}
Pretrained LM has shown tremendous success in text generation, and many have studied large autoregressive LMs such as GPT-2 on conditional generation by performing operations on the sequence space of the language models. For example, \citet{Dathathri2020Plug} proposes a plug-and-play framework that utilizes gradients of attribute classifiers to modify the hidden states of the pretrained LM at every step, named PPLM. 
FUDGE~\cite{DBLP:journals/corr/abs-2104-05218} follows a similar architecture but incorporates classifiers that predict the conditional probability of a complete sentence given prefixes to adjust the vocabulary probability distribution given by LM. Differing from these two approaches with left-to-right decoding, MUCOCO~\cite{DBLP:conf/nips/KumarMST21} formulates the decoding process as a multi-objective continuous optimization that combines loss of pretrained LM and attributes classifiers. The optimization gradient is applied directly to the soft representation consisting of each token's vocabulary distribution. COLD~\cite{QinCOLD} adopts the exact soft representation but uses an energy-based model with attribute constraints and Langevin Dynamics to sample. 
\begin{table}[t]
\setlength\tabcolsep{5.5pt}
    \scriptsize
    \centering
    \vspace{-15pt}
    \begin{tabular}{lccccc}
    \toprule
    \multirow{3}{*}{Methods} & \multicolumn{2}{c}{Accuracy$\uparrow$} & \multicolumn{2}{c}{Content$\uparrow$}  & Fluency$\downarrow$\\\cmidrule(r){2-3}\cmidrule(r){4-5}\cmidrule(r){6-6}
    & Sentiment & Tense & iBL  &  CTC &PPL   \\
    \midrule
     FUDGE & 0.36  & 0.56  & \textbf{56.5}  & 0.450 & \textbf{17.3}\\
    Ours  & \textbf{0.95}  & \textbf{0.95}  & {37.1}& \textbf{0.465} & 30.1 \\\bottomrule
    \end{tabular}
    \caption{Automatic evaluation results of text editing with compositional attributes on Yelp review dataset.}
    \label{tab:tst_multiple}
    \vspace{-10pt}
\end{table}
\subsection{Text Control in Latent Space}
Another common approach to control text generation is modifying text representation in the latent space. Some methods~\cite{DBLP:conf/icml/MuellerGJ17, DBLP:conf/aaai/LiuFZPL20} utilize a VAE to encode the input sequence into $\bm z$ in the latent space and then use attribute networks that are jointly trained with the VAE to obtain $\bm z^\prime$ that can be decoded into the desired sequence. 
PPVAE~\cite{DBLP:conf/acl/DuanXPHL20} uses an unconditional Pre-train VAE and a conditional Plugin-VAE to achieve the goal.
Plug and Play \cite{mai-etal-2020-plug} follows a similar framework but replaces the VAE with an Auto-encoder and the Plugin-VAE with an MLP to obtain a desired vector $\bm z^\prime$. Some methods use an attribute classifier to edit the latent representation $\bm{z}$ with Fast-Gradient-Iterative-Modification \cite{DBLP:conf/nips/WangH019}. Because of the recent success of diffusion models, LDEBM \cite{yulatent} proposes a diffusion process in the latent space whose reverse process is constructed with a sequence of EBMs for text generation. 

\section{Conclusions}
 We have developed a new efficient approach that performs composable control operations in the compact latent space of text, named \textsc{LatentOps}. 
The proposed method permits combining arbitrary operators applied on a latent vector, resulting in an energy-based distribution on the low-dimensional continuous latent space.
We develop an efficient and robust sampler based on ODEs that effectively samples from the distribution guided by gradients.
We connect the latent space to popular pretrained LM by efficient adaptation without finetuning the whole model.
We showcase its compositionality, flexibility and firm performance on several distinct tasks.
In future work, we can explore the control of more complicated texts.

\section*{Ethical Considerations}
The contributions of this paper mostly focus around the fundamental challenges in designing an efficient approach for composable text operations in the compact latent space of text, and the proposed method is examined on commonly used public datasets. This work has applications in conditional text generation, text style transfer, data augmentation, and few-shot learning.

VAEs, the framework of our latent model, are trained to mimic the training data distribution, and , bias introduced in data collection will make VAEs generate samples with a similar bias. Additional bias could be introduced during model design or training. However, such techniques could be misused to produce fake or misleading information, and researchers should be aware of these risks and explore the techniques responsibly.
\section*{Limitations}
The primary focus of this paper is the analysis of single sentences. Our objective has been to deeply understand the potential of the proposed method, with a particular emphasis on its controllability and compositionality. Analyzing single sentences offers a relatively controlled setting, making it easier to derive clear insights and manage data complexities.

However, it's important to recognize that our findings, while based on single sentences, may not directly translate to longer textual content. Lengthier texts bring with them complex structures, dependencies, and nuanced contexts that might affect the performance of our methods. Adapting to these challenges may require further refinements.


Considering the scope of our current investigation, there exists significant opportunity for future research. This includes not only adapting our methodology to handle more intricate textual scenarios but also contrasting its performance with other potential approaches. Such explorations remain promising directions for forthcoming studies.
\bibliography{anthology}

\begin{thebibliography}{64}
\expandafter\ifx\csname natexlab\endcsname\relax\def\natexlab#1{#1}\fi

\bibitem[{Anderson(1982)}]{anderson1982reverse}
Brian~DO Anderson. 1982.
\newblock Reverse-time diffusion equation models.
\newblock \emph{Stochastic Processes and their Applications}, 12(3):313--326.

\bibitem[{Bird et~al.(2009)Bird, Klein, and Loper}]{DBLP:books/daglib/0022921}
Steven Bird, Ewan Klein, and Edward Loper. 2009.
\newblock \href {http://www.oreilly.de/catalog/9780596516499/index.html}
  {\emph{Natural Language Processing with Python}}.
\newblock O'Reilly.

\bibitem[{Bishop and Nasrabadi(2006)}]{bishop2006pattern}
Christopher~M Bishop and Nasser~M Nasrabadi. 2006.
\newblock \emph{Pattern recognition and machine learning}, volume~4.
\newblock Springer.

\bibitem[{Bowman et~al.(2016)Bowman, Vilnis, Vinyals, Dai, J{\'{o}}zefowicz,
  and Bengio}]{DBLP:conf/conll/BowmanVVDJB16}
Samuel~R. Bowman, Luke Vilnis, Oriol Vinyals, Andrew~M. Dai, Rafal
  J{\'{o}}zefowicz, and Samy Bengio. 2016.
\newblock \href {https://doi.org/10.18653/v1/k16-1002} {Generating sentences
  from a continuous space}.
\newblock In \emph{Proceedings of the 20th {SIGNLL} Conference on Computational
  Natural Language Learning, CoNLL 2016, Berlin, Germany, August 11-12, 2016},
  pages 10--21. {ACL}.

\bibitem[{Burrage et~al.(2000)Burrage, Burrage, and
  Mitsui}]{burrage2000numerical}
Kevin Burrage, Pamela Burrage, and Taketomo Mitsui. 2000.
\newblock Numerical solutions of stochastic differential
  equations--implementation and stability issues.
\newblock \emph{Journal of computational and applied mathematics},
  125(1-2):171--182.

\bibitem[{Calvo et~al.(1990)Calvo, Montijano, and Randez}]{calvo1990fifth}
M~Calvo, JI~Montijano, and L~Randez. 1990.
\newblock A fifth-order interpolant for the dormand and prince runge-kutta
  method.
\newblock \emph{Journal of computational and applied mathematics},
  29(1):91--100.

\bibitem[{Chen et~al.(2021)Chen, Amos, and Nickel}]{chen2021eventfn}
Ricky T.~Q. Chen, Brandon Amos, and Maximilian Nickel. 2021.
\newblock Learning neural event functions for ordinary differential equations.
\newblock \emph{International Conference on Learning Representations}.

\bibitem[{Chen et~al.(2018)Chen, Rubanova, Bettencourt, and
  Duvenaud}]{chen2018neuralode}
Ricky T.~Q. Chen, Yulia Rubanova, Jesse Bettencourt, and David Duvenaud. 2018.
\newblock Neural ordinary differential equations.
\newblock \emph{Advances in Neural Information Processing Systems}.

\bibitem[{Dai et~al.(2019)Dai, Liang, Qiu, and Huang}]{DBLP:conf/acl/DaiLQH19}
Ning Dai, Jianze Liang, Xipeng Qiu, and Xuanjing Huang. 2019.
\newblock \href {https://doi.org/10.18653/v1/p19-1601} {Style transformer:
  Unpaired text style transfer without disentangled latent representation}.
\newblock In \emph{Proceedings of the 57th Conference of the Association for
  Computational Linguistics, {ACL} 2019, Florence, Italy, July 28- August 2,
  2019, Volume 1: Long Papers}, pages 5997--6007. Association for Computational
  Linguistics.

\bibitem[{Dathathri et~al.(2020)Dathathri, Madotto, Lan, Hung, Frank, Molino,
  Yosinski, and Liu}]{Dathathri2020Plug}
Sumanth Dathathri, Andrea Madotto, Janice Lan, Jane Hung, Eric Frank, Piero
  Molino, Jason Yosinski, and Rosanne Liu. 2020.
\newblock \href {https://openreview.net/forum?id=H1edEyBKDS} {Plug and play
  language models: A simple approach to controlled text generation}.
\newblock In \emph{International Conference on Learning Representations}.

\bibitem[{Deng et~al.(2021)Deng, Tan, Liu, Xing, and Hu}]{ctc_score}
Mingkai Deng, Bowen Tan, Zhengzhong Liu, Eric~P. Xing, and Zhiting Hu. 2021.
\newblock \href {https://doi.org/10.18653/v1/2021.emnlp-main.599} {Compression,
  transduction, and creation: {A} unified framework for evaluating natural
  language generation}.
\newblock In \emph{Proceedings of the 2021 Conference on Empirical Methods in
  Natural Language Processing, {EMNLP} 2021, Virtual Event / Punta Cana,
  Dominican Republic, 7-11 November, 2021}, pages 7580--7605. Association for
  Computational Linguistics.

\bibitem[{Deng et~al.(2020)Deng, Bakhtin, Ott, Szlam, and
  Ranzato}]{DBLP:conf/iclr/DengBOSR20}
Yuntian Deng, Anton Bakhtin, Myle Ott, Arthur Szlam, and Marc'Aurelio Ranzato.
  2020.
\newblock \href {https://openreview.net/forum?id=B1l4SgHKDH} {Residual
  energy-based models for text generation}.
\newblock In \emph{8th International Conference on Learning Representations,
  {ICLR} 2020, Addis Ababa, Ethiopia, April 26-30, 2020}. OpenReview.net.

\bibitem[{Devlin et~al.(2019)Devlin, Chang, Lee, and Toutanova}]{bert}
Jacob Devlin, Ming{-}Wei Chang, Kenton Lee, and Kristina Toutanova. 2019.
\newblock \href {https://doi.org/10.18653/v1/n19-1423} {{BERT:} pre-training of
  deep bidirectional transformers for language understanding}.
\newblock In \emph{Proceedings of the 2019 Conference of the North American
  Chapter of the Association for Computational Linguistics: Human Language
  Technologies, {NAACL-HLT} 2019, Minneapolis, MN, USA, June 2-7, 2019, Volume
  1 (Long and Short Papers)}, pages 4171--4186. Association for Computational
  Linguistics.

\bibitem[{Du et~al.(2020)Du, Li, and
  Mordatch}]{DBLP:journals/corr/abs-2004-06030}
Yilun Du, Shuang Li, and Igor Mordatch. 2020.
\newblock \href {http://arxiv.org/abs/2004.06030} {Compositional visual
  generation and inference with energy based models}.
\newblock \emph{CoRR}, abs/2004.06030.

\bibitem[{Du and
  Mordatch(2019{\natexlab{a}})}]{DBLP:journals/corr/abs-1903-08689}
Yilun Du and Igor Mordatch. 2019{\natexlab{a}}.
\newblock \href {http://arxiv.org/abs/1903.08689} {Implicit generation and
  generalization in energy-based models}.
\newblock \emph{CoRR}, abs/1903.08689.

\bibitem[{Du and Mordatch(2019{\natexlab{b}})}]{DBLP:conf/nips/DuM19}
Yilun Du and Igor Mordatch. 2019{\natexlab{b}}.
\newblock \href
  {https://proceedings.neurips.cc/paper/2019/hash/378a063b8fdb1db941e34f4bde584c7d-Abstract.html}
  {Implicit generation and modeling with energy based models}.
\newblock In \emph{Advances in Neural Information Processing Systems 32: Annual
  Conference on Neural Information Processing Systems 2019, NeurIPS 2019,
  December 8-14, 2019, Vancouver, BC, Canada}, pages 3603--3613.

\bibitem[{Duan et~al.(2020)Duan, Xu, Pei, Han, and
  Li}]{DBLP:conf/acl/DuanXPHL20}
Yu~Duan, Canwen Xu, Jiaxin Pei, Jialong Han, and Chenliang Li. 2020.
\newblock \href {https://doi.org/10.18653/v1/2020.acl-main.23} {Pre-train and
  plug-in: Flexible conditional text generation with variational
  auto-encoders}.
\newblock In \emph{Proceedings of the 58th Annual Meeting of the Association
  for Computational Linguistics, {ACL} 2020, Online, July 5-10, 2020}, pages
  253--262. Association for Computational Linguistics.

\bibitem[{Engstler and Lubich(1997)}]{engstler1997mur8}
Chr Engstler and Chr Lubich. 1997.
\newblock Mur8: a multirate extension of the eighth-order dormand-prince
  method.
\newblock \emph{Applied numerical mathematics}, 25(2-3):185--192.

\bibitem[{Euler(1824)}]{euler1824institutionum}
Leonhard Euler. 1824.
\newblock \emph{Institutionum calculi integralis}, volume~1.
\newblock impensis Academiae imperialis scientiarum.

\bibitem[{Goodfellow et~al.(2014)Goodfellow, Pouget-Abadie, Mirza, Xu,
  Warde-Farley, Ozair, Courville, and Bengio}]{goodfellow2014generative}
Ian Goodfellow, Jean Pouget-Abadie, Mehdi Mirza, Bing Xu, David Warde-Farley,
  Sherjil Ozair, Aaron Courville, and Yoshua Bengio. 2014.
\newblock Generative adversarial nets.
\newblock \emph{Advances in neural information processing systems}, 27.

\bibitem[{Grathwohl et~al.(2020)Grathwohl, Wang, Jacobsen, Duvenaud, Norouzi,
  and Swersky}]{DBLP:conf/iclr/GrathwohlWJD0S20}
Will Grathwohl, Kuan{-}Chieh Wang, J{\"{o}}rn{-}Henrik Jacobsen, David
  Duvenaud, Mohammad Norouzi, and Kevin Swersky. 2020.
\newblock \href {https://openreview.net/forum?id=Hkxzx0NtDB} {Your classifier
  is secretly an energy based model and you should treat it like one}.
\newblock In \emph{8th International Conference on Learning Representations,
  {ICLR} 2020, Addis Ababa, Ethiopia, April 26-30, 2020}. OpenReview.net.

\bibitem[{He and McAuley(2016)}]{DBLP:conf/www/HeM16}
Ruining He and Julian~J. McAuley. 2016.
\newblock \href {https://doi.org/10.1145/2872427.2883037} {Ups and downs:
  Modeling the visual evolution of fashion trends with one-class collaborative
  filtering}.
\newblock In \emph{Proceedings of the 25th International Conference on World
  Wide Web, {WWW} 2016, Montreal, Canada, April 11 - 15, 2016}, pages 507--517.
  {ACM}.

\bibitem[{Higham(2001)}]{higham2001algorithmic}
Desmond~J Higham. 2001.
\newblock An algorithmic introduction to numerical simulation of stochastic
  differential equations.
\newblock \emph{SIAM review}, 43(3):525--546.

\bibitem[{Hu and Li(2021)}]{hu2021causal}
Zhiting Hu and Li~Erran Li. 2021.
\newblock A causal lens for controllable text generation.
\newblock \emph{Advances in Neural Information Processing Systems},
  34:24941--24955.

\bibitem[{Hu et~al.(2017)Hu, Yang, Liang, Salakhutdinov, and
  Xing}]{hu2017toward}
Zhiting Hu, Zichao Yang, Xiaodan Liang, Ruslan Salakhutdinov, and Eric~P Xing.
  2017.
\newblock Toward controlled generation of text.
\newblock In \emph{International conference on machine learning}, pages
  1587--1596. PMLR.

\bibitem[{Hu et~al.(2018)Hu, Yang, Salakhutdinov, Qin, Liang, Dong, and
  Xing}]{DBLP:conf/nips/HuYSQLDX18}
Zhiting Hu, Zichao Yang, Ruslan Salakhutdinov, Lianhui Qin, Xiaodan Liang,
  Haoye Dong, and Eric~P. Xing. 2018.
\newblock \href
  {https://proceedings.neurips.cc/paper/2018/hash/d7e77c835af3d2a803c1cf28d60575bc-Abstract.html}
  {Deep generative models with learnable knowledge constraints}.
\newblock In \emph{Advances in Neural Information Processing Systems 31: Annual
  Conference on Neural Information Processing Systems 2018, NeurIPS 2018,
  December 3-8, 2018, Montr{\'{e}}al, Canada}, pages 10522--10533.

\bibitem[{Keskar et~al.(2019)Keskar, McCann, Varshney, Xiong, and
  Socher}]{keskarCTRL2019}
Nitish~Shirish Keskar, Bryan McCann, Lav Varshney, Caiming Xiong, and Richard
  Socher. 2019.
\newblock {CTRL - A Conditional Transformer Language Model for Controllable
  Generation}.
\newblock \emph{arXiv preprint arXiv:1909.05858}.

\bibitem[{Khalifa et~al.(2021)Khalifa, Elsahar, and
  Dymetman}]{DBLP:conf/iclr/KhalifaED21}
Muhammad Khalifa, Hady Elsahar, and Marc Dymetman. 2021.
\newblock \href {https://openreview.net/forum?id=jWkw45-9AbL} {A distributional
  approach to controlled text generation}.
\newblock In \emph{9th International Conference on Learning Representations,
  {ICLR} 2021, Virtual Event, Austria, May 3-7, 2021}. OpenReview.net.

\bibitem[{Kingma and Welling(2014)}]{DBLP:journals/corr/KingmaW13}
Diederik~P. Kingma and Max Welling. 2014.
\newblock \href {http://arxiv.org/abs/1312.6114} {Auto-encoding variational
  bayes}.
\newblock In \emph{2nd International Conference on Learning Representations,
  {ICLR} 2014, Banff, AB, Canada, April 14-16, 2014, Conference Track
  Proceedings}.

\bibitem[{Krause et~al.(2021)Krause, Gotmare, McCann, Keskar, Joty, Socher, and
  Rajani}]{DBLP:conf/emnlp/KrauseGMKJSR21}
Ben Krause, Akhilesh~Deepak Gotmare, Bryan McCann, Nitish~Shirish Keskar,
  Shafiq~R. Joty, Richard Socher, and Nazneen~Fatema Rajani. 2021.
\newblock \href {https://doi.org/10.18653/v1/2021.findings-emnlp.424} {Gedi:
  Generative discriminator guided sequence generation}.
\newblock In \emph{Findings of the Association for Computational Linguistics:
  {EMNLP} 2021, Virtual Event / Punta Cana, Dominican Republic, 16-20 November,
  2021}, pages 4929--4952. Association for Computational Linguistics.

\bibitem[{Kumar et~al.(2021)Kumar, Malmi, Severyn, and
  Tsvetkov}]{DBLP:conf/nips/KumarMST21}
Sachin Kumar, Eric Malmi, Aliaksei Severyn, and Yulia Tsvetkov. 2021.
\newblock \href
  {https://proceedings.neurips.cc/paper/2021/hash/79ec2a4246feb2126ecf43c4a4418002-Abstract.html}
  {Controlled text generation as continuous optimization with multiple
  constraints}.
\newblock In \emph{Advances in Neural Information Processing Systems 34: Annual
  Conference on Neural Information Processing Systems 2021, NeurIPS 2021,
  December 6-14, 2021, virtual}, pages 14542--14554.

\bibitem[{Li et~al.(2020)Li, Gao, Li, Peng, Li, Zhang, and
  Gao}]{li-etal-2020-optimus}
Chunyuan Li, Xiang Gao, Yuan Li, Baolin Peng, Xiujun Li, Yizhe Zhang, and
  Jianfeng Gao. 2020.
\newblock \href {https://doi.org/10.18653/v1/2020.emnlp-main.378} {Optimus:
  Organizing sentences via pre-trained modeling of a latent space}.
\newblock In \emph{Proceedings of the 2020 Conference on Empirical Methods in
  Natural Language Processing (EMNLP)}, pages 4678--4699, Online. Association
  for Computational Linguistics.

\bibitem[{Li et~al.(2018)Li, Jia, He, and Liang}]{DBLP:conf/naacl/LiJHL18}
Juncen Li, Robin Jia, He~He, and Percy Liang. 2018.
\newblock \href {https://doi.org/10.18653/v1/n18-1169} {Delete, retrieve,
  generate: a simple approach to sentiment and style transfer}.
\newblock In \emph{Proceedings of the 2018 Conference of the North American
  Chapter of the Association for Computational Linguistics: Human Language
  Technologies, {NAACL-HLT} 2018, New Orleans, Louisiana, USA, June 1-6, 2018,
  Volume 1 (Long Papers)}, pages 1865--1874. Association for Computational
  Linguistics.

\bibitem[{Li et~al.(2022)Li, Thickstun, Gulrajani, Liang, and
  Hashimoto}]{li2022diffusion}
Xiang~Lisa Li, John Thickstun, Ishaan Gulrajani, Percy Liang, and Tatsunori~B
  Hashimoto. 2022.
\newblock {Diffusion-LM} improves controllable text generation.
\newblock \emph{arXiv preprint arXiv:2205.14217}.

\bibitem[{Liu et~al.(2020)Liu, Fu, Zhang, Pal, and
  Lv}]{DBLP:conf/aaai/LiuFZPL20}
Dayiheng Liu, Jie Fu, Yidan Zhang, Chris Pal, and Jiancheng Lv. 2020.
\newblock \href {https://ojs.aaai.org/index.php/AAAI/article/view/6355}
  {Revision in continuous space: Unsupervised text style transfer without
  adversarial learning}.
\newblock In \emph{The Thirty-Fourth {AAAI} Conference on Artificial
  Intelligence, {AAAI} 2020, The Thirty-Second Innovative Applications of
  Artificial Intelligence Conference, {IAAI} 2020, The Tenth {AAAI} Symposium
  on Educational Advances in Artificial Intelligence, {EAAI} 2020, New York,
  NY, USA, February 7-12, 2020}, pages 8376--8383. {AAAI} Press.

\bibitem[{Liu et~al.(2021{\natexlab{a}})Liu, Yang, Tao, Liang, Li, Zhou, Cui,
  and Hu}]{DBLP:journals/corr/abs-2106-15078}
Guangyi Liu, Zichao Yang, Tianhua Tao, Xiaodan Liang, Zhen Li, Bowen Zhou,
  Shuguang Cui, and Zhiting Hu. 2021{\natexlab{a}}.
\newblock \href {http://arxiv.org/abs/2106.15078} {Don't take it literally: An
  edit-invariant sequence loss for text generation}.
\newblock \emph{CoRR}, abs/2106.15078.

\bibitem[{Liu et~al.(2021{\natexlab{b}})Liu, Neubig, and
  Wieting}]{DBLP:conf/naacl/LiuNW21}
Yixin Liu, Graham Neubig, and John Wieting. 2021{\natexlab{b}}.
\newblock \href {https://doi.org/10.18653/v1/2021.naacl-main.337} {On learning
  text style transfer with direct rewards}.
\newblock In \emph{Proceedings of the 2021 Conference of the North American
  Chapter of the Association for Computational Linguistics: Human Language
  Technologies, {NAACL-HLT} 2021, Online, June 6-11, 2021}, pages 4262--4273.
  Association for Computational Linguistics.

\bibitem[{Ma et~al.(2018)Ma, Chen, Jin, Flammarion, and
  Jordan}]{DBLP:journals/corr/abs-1811-08413}
Yi{-}An Ma, Yuansi Chen, Chi Jin, Nicolas Flammarion, and Michael~I. Jordan.
  2018.
\newblock \href {http://arxiv.org/abs/1811.08413} {Sampling can be faster than
  optimization}.
\newblock \emph{CoRR}, abs/1811.08413.

\bibitem[{Madaan et~al.(2020)Madaan, Setlur, Parekh, P{\'{o}}czos, Neubig,
  Yang, Salakhutdinov, Black, and Prabhumoye}]{DBLP:conf/acl/MadaanSPPNYSBP20}
Aman Madaan, Amrith Setlur, Tanmay Parekh, Barnab{\'{a}}s P{\'{o}}czos, Graham
  Neubig, Yiming Yang, Ruslan Salakhutdinov, Alan~W. Black, and Shrimai
  Prabhumoye. 2020.
\newblock \href {https://doi.org/10.18653/v1/2020.acl-main.169} {Politeness
  transfer: {A} tag and generate approach}.
\newblock In \emph{Proceedings of the 58th Annual Meeting of the Association
  for Computational Linguistics, {ACL} 2020, Online, July 5-10, 2020}, pages
  1869--1881. Association for Computational Linguistics.

\bibitem[{Mai et~al.(2020{\natexlab{a}})Mai, Pappas, Montero, Smith, and
  Henderson}]{DBLP:conf/emnlp/MaiPMSH20}
Florian Mai, Nikolaos Pappas, Ivan Montero, Noah~A. Smith, and James Henderson.
  2020{\natexlab{a}}.
\newblock \href {https://doi.org/10.18653/v1/2020.emnlp-main.491} {Plug and
  play autoencoders for conditional text generation}.
\newblock In \emph{Proceedings of the 2020 Conference on Empirical Methods in
  Natural Language Processing, {EMNLP} 2020, Online, November 16-20, 2020},
  pages 6076--6092. Association for Computational Linguistics.

\bibitem[{Mai et~al.(2020{\natexlab{b}})Mai, Pappas, Montero, Smith, and
  Henderson}]{mai-etal-2020-plug}
Florian Mai, Nikolaos Pappas, Ivan Montero, Noah~A. Smith, and James Henderson.
  2020{\natexlab{b}}.
\newblock \href {https://doi.org/10.18653/v1/2020.emnlp-main.491} {Plug and
  play autoencoders for conditional text generation}.
\newblock In \emph{Proceedings of the 2020 Conference on Empirical Methods in
  Natural Language Processing (EMNLP)}, pages 6076--6092, Online. Association
  for Computational Linguistics.

\bibitem[{Maoutsa et~al.(2020)Maoutsa, Reich, and
  Opper}]{DBLP:journals/entropy/MaoutsaRO20}
Dimitra Maoutsa, Sebastian Reich, and Manfred Opper. 2020.
\newblock \href {https://doi.org/10.3390/e22080802} {Interacting particle
  solutions of fokker-planck equations through gradient-log-density
  estimation}.
\newblock \emph{Entropy}, 22(8):802.

\bibitem[{Mireshghallah et~al.(2022)Mireshghallah, Goyal, and
  Berg{-}Kirkpatrick}]{DBLP:journals/corr/abs-2203-13299}
Fatemehsadat Mireshghallah, Kartik Goyal, and Taylor Berg{-}Kirkpatrick. 2022.
\newblock \href {https://doi.org/10.48550/arXiv.2203.13299} {Mix and match:
  Learning-free controllable text generation using energy language models}.
\newblock \emph{CoRR}, abs/2203.13299.

\bibitem[{Mueller et~al.(2017)Mueller, Gifford, and
  Jaakkola}]{DBLP:conf/icml/MuellerGJ17}
Jonas Mueller, David~K. Gifford, and Tommi~S. Jaakkola. 2017.
\newblock \href {http://proceedings.mlr.press/v70/mueller17a.html} {Sequence to
  better sequence: Continuous revision of combinatorial structures}.
\newblock In \emph{Proceedings of the 34th International Conference on Machine
  Learning, {ICML} 2017, Sydney, NSW, Australia, 6-11 August 2017}, volume~70
  of \emph{Proceedings of Machine Learning Research}, pages 2536--2544. {PMLR}.

\bibitem[{Nie et~al.(2021)Nie, Vahdat, and Anandkumar}]{nie2021controllable}
Weili Nie, Arash Vahdat, and Anima Anandkumar. 2021.
\newblock \href
  {https://proceedings.neurips.cc/paper/2021/hash/701d804549a4a23d3cae801dac6c2c75-Abstract.html}
  {Controllable and compositional generation with latent-space energy-based
  models}.
\newblock In \emph{Advances in Neural Information Processing Systems 34: Annual
  Conference on Neural Information Processing Systems 2021, NeurIPS 2021,
  December 6-14, 2021, virtual}, pages 13497--13510.

\bibitem[{Pillutla et~al.(2021)Pillutla, Swayamdipta, Zellers, Thickstun,
  Welleck, Choi, and Harchaoui}]{pillutla2021mauve}
Krishna Pillutla, Swabha Swayamdipta, Rowan Zellers, John Thickstun, Sean
  Welleck, Yejin Choi, and Zaid Harchaoui. 2021.
\newblock Mauve: Measuring the gap between neural text and human text using
  divergence frontiers.
\newblock \emph{Advances in Neural Information Processing Systems},
  34:4816--4828.

\bibitem[{Qin et~al.(2020)Qin, Shwartz, West, Bhagavatula, Hwang, Le~Bras,
  Bosselut, and Choi}]{qin2020back}
Lianhui Qin, Vered Shwartz, Peter West, Chandra Bhagavatula, Jena~D Hwang,
  Ronan Le~Bras, Antoine Bosselut, and Yejin Choi. 2020.
\newblock Back to the future: Unsupervised backprop-based decoding for
  counterfactual and abductive commonsense reasoning.
\newblock In \emph{Proceedings of the 2020 Conference on Empirical Methods in
  Natural Language Processing (EMNLP)}, pages 794--805.

\bibitem[{Qin et~al.(2022)Qin, Welleck, Khashabi, and Choi}]{QinCOLD}
Lianhui Qin, Sean Welleck, Daniel Khashabi, and Yejin Choi. 2022.
\newblock \href {http://arxiv.org/abs/arXiv:2202.11705} {Cold decoding:
  Energy-based constrained text generation with langevin dynamics}.

\bibitem[{Radford et~al.(2019)Radford, Wu, Child, Luan, Amodei, Sutskever
  et~al.}]{gpt2}
Alec Radford, Jeffrey Wu, Rewon Child, David Luan, Dario Amodei, Ilya
  Sutskever, et~al. 2019.
\newblock Language models are unsupervised multitask learners.
\newblock \emph{OpenAI blog}, 1(8):9.

\bibitem[{Rao and Tetreault(2018)}]{DBLP:conf/naacl/RaoT18}
Sudha Rao and Joel~R. Tetreault. 2018.
\newblock \href {https://doi.org/10.18653/v1/n18-1012} {Dear sir or madam, may
  {I} introduce the {GYAFC} dataset: Corpus, benchmarks and metrics for
  formality style transfer}.
\newblock In \emph{Proceedings of the 2018 Conference of the North American
  Chapter of the Association for Computational Linguistics: Human Language
  Technologies, {NAACL-HLT} 2018, New Orleans, Louisiana, USA, June 1-6, 2018,
  Volume 1 (Long Papers)}, pages 129--140. Association for Computational
  Linguistics.

\bibitem[{Rezende et~al.(2014)Rezende, Mohamed, and
  Wierstra}]{DBLP:conf/icml/RezendeMW14}
Danilo~Jimenez Rezende, Shakir Mohamed, and Daan Wierstra. 2014.
\newblock \href {http://proceedings.mlr.press/v32/rezende14.html} {Stochastic
  backpropagation and approximate inference in deep generative models}.
\newblock In \emph{Proceedings of the 31th International Conference on Machine
  Learning, {ICML} 2014, Beijing, China, 21-26 June 2014}, volume~32 of
  \emph{{JMLR} Workshop and Conference Proceedings}, pages 1278--1286.
  JMLR.org.

\bibitem[{R{\"o}{\ss}ler(2009)}]{rossler2009second}
Andreas R{\"o}{\ss}ler. 2009.
\newblock Second order runge--kutta methods for it{\^o} stochastic differential
  equations.
\newblock \emph{SIAM Journal on Numerical Analysis}, 47(3):1713--1738.

\bibitem[{Shen et~al.(2017)Shen, Lei, Barzilay, and
  Jaakkola}]{DBLP:conf/nips/ShenLBJ17}
Tianxiao Shen, Tao Lei, Regina Barzilay, and Tommi~S. Jaakkola. 2017.
\newblock \href
  {https://proceedings.neurips.cc/paper/2017/hash/2d2c8394e31101a261abf1784302bf75-Abstract.html}
  {Style transfer from non-parallel text by cross-alignment}.
\newblock In \emph{Advances in Neural Information Processing Systems 30: Annual
  Conference on Neural Information Processing Systems 2017, December 4-9, 2017,
  Long Beach, CA, {USA}}, pages 6830--6841.

\bibitem[{Shen et~al.(2020)Shen, Mueller, Barzilay, and
  Jaakkola}]{DBLP:conf/icml/ShenMBJ20}
Tianxiao Shen, Jonas Mueller, Regina Barzilay, and Tommi~S. Jaakkola. 2020.
\newblock \href {http://proceedings.mlr.press/v119/shen20c.html} {Educating
  text autoencoders: Latent representation guidance via denoising}.
\newblock In \emph{Proceedings of the 37th International Conference on Machine
  Learning, {ICML} 2020, 13-18 July 2020, Virtual Event}, volume 119 of
  \emph{Proceedings of Machine Learning Research}, pages 8719--8729. {PMLR}.

\bibitem[{Song and Ermon(2019)}]{DBLP:conf/nips/SongE19}
Yang Song and Stefano Ermon. 2019.
\newblock \href
  {https://proceedings.neurips.cc/paper/2019/hash/3001ef257407d5a371a96dcd947c7d93-Abstract.html}
  {Generative modeling by estimating gradients of the data distribution}.
\newblock In \emph{Advances in Neural Information Processing Systems 32: Annual
  Conference on Neural Information Processing Systems 2019, NeurIPS 2019,
  December 8-14, 2019, Vancouver, BC, Canada}, pages 11895--11907.

\bibitem[{Song et~al.(2021)Song, Sohl{-}Dickstein, Kingma, Kumar, Ermon, and
  Poole}]{DBLP:conf/iclr/0011SKKEP21}
Yang Song, Jascha Sohl{-}Dickstein, Diederik~P. Kingma, Abhishek Kumar, Stefano
  Ermon, and Ben Poole. 2021.
\newblock \href {https://openreview.net/forum?id=PxTIG12RRHS} {Score-based
  generative modeling through stochastic differential equations}.
\newblock In \emph{9th International Conference on Learning Representations,
  {ICLR} 2021, Virtual Event, Austria, May 3-7, 2021}. OpenReview.net.

\bibitem[{Sudhakar et~al.(2019)Sudhakar, Upadhyay, and
  Maheswaran}]{DBLP:conf/emnlp/SudhakarUM19}
Akhilesh Sudhakar, Bhargav Upadhyay, and Arjun Maheswaran. 2019.
\newblock \href {https://doi.org/10.18653/v1/D19-1322} {"transforming" delete,
  retrieve, generate approach for controlled text style transfer}.
\newblock In \emph{Proceedings of the 2019 Conference on Empirical Methods in
  Natural Language Processing and the 9th International Joint Conference on
  Natural Language Processing, {EMNLP-IJCNLP} 2019, Hong Kong, China, November
  3-7, 2019}, pages 3267--3277. Association for Computational Linguistics.

\bibitem[{Turc et~al.(2019)Turc, Chang, Lee, and
  Toutanova}]{DBLP:journals/corr/abs-1908-08962}
Iulia Turc, Ming{-}Wei Chang, Kenton Lee, and Kristina Toutanova. 2019.
\newblock \href {http://arxiv.org/abs/1908.08962} {Well-read students learn
  better: The impact of student initialization on knowledge distillation}.
\newblock \emph{CoRR}, abs/1908.08962.

\bibitem[{Vahdat et~al.(2021)Vahdat, Kreis, and Kautz}]{Vahdat_LSGM}
Arash Vahdat, Karsten Kreis, and Jan Kautz. 2021.
\newblock \href
  {https://proceedings.neurips.cc/paper/2021/file/5dca4c6b9e244d24a30b4c45601d9720-Paper.pdf}
  {Score-based generative modeling in latent space}.
\newblock In \emph{Advances in Neural Information Processing Systems},
  volume~34, pages 11287--11302. Curran Associates, Inc.

\bibitem[{Wang et~al.(2019)Wang, Hua, and Wan}]{DBLP:conf/nips/WangH019}
Ke~Wang, Hang Hua, and Xiaojun Wan. 2019.
\newblock \href
  {https://proceedings.neurips.cc/paper/2019/hash/8804f94e16ba5b680e239a554a08f7d2-Abstract.html}
  {Controllable unsupervised text attribute transfer via editing entangled
  latent representation}.
\newblock In \emph{Advances in Neural Information Processing Systems 32: Annual
  Conference on Neural Information Processing Systems 2019, NeurIPS 2019,
  December 8-14, 2019, Vancouver, BC, Canada}, pages 11034--11044.

\bibitem[{Welling and Teh(2011)}]{DBLP:conf/icml/WellingT11}
Max Welling and Yee~Whye Teh. 2011.
\newblock \href {https://icml.cc/2011/papers/398\_icmlpaper.pdf} {Bayesian
  learning via stochastic gradient langevin dynamics}.
\newblock In \emph{Proceedings of the 28th International Conference on Machine
  Learning, {ICML} 2011, Bellevue, Washington, USA, June 28 - July 2, 2011},
  pages 681--688. Omnipress.

\bibitem[{Yang and Klein(2021)}]{DBLP:journals/corr/abs-2104-05218}
Kevin Yang and Dan Klein. 2021.
\newblock \href {http://arxiv.org/abs/2104.05218} {{FUDGE:} controlled text
  generation with future discriminators}.
\newblock \emph{CoRR}, abs/2104.05218.

\bibitem[{Yu et~al.(2022)Yu, Xie, Ma, Jia, Pang, Gao, Zhu, Zhu, and
  Wu}]{yulatent}
Peiyu Yu, Sirui Xie, Xiaojian Ma, Baoxiong Jia, Bo~Pang, Ruiqi Gao, Yixin Zhu,
  Song{-}Chun Zhu, and Ying~Nian Wu. 2022.
\newblock \href {https://doi.org/10.48550/arXiv.2206.05895} {Latent diffusion
  energy-based model for interpretable text modeling}.
\newblock \emph{CoRR}, abs/2206.05895.

\bibitem[{Ziegler et~al.(2019)Ziegler, Stiennon, Wu, Brown, Radford, Amodei,
  Christiano, and Irving}]{DBLP:journals/corr/abs-1909-08593}
Daniel~M. Ziegler, Nisan Stiennon, Jeffrey Wu, Tom~B. Brown, Alec Radford,
  Dario Amodei, Paul~F. Christiano, and Geoffrey Irving. 2019.
\newblock \href {http://arxiv.org/abs/1909.08593} {Fine-tuning language models
  from human preferences}.
\newblock \emph{CoRR}, abs/1909.08593.

\end{thebibliography}
\bibliographystyle{acl_natbib}

\clearpage
\appendix

\onecolumn
\section{Derivation of ODE Formulation}
\label{app:derivation_ode}
\newcommand{\td}{\text d}
\newcommand{\gxt}{\bm{G}(\bm{x},t)}
\newcommand{\bmx}{\bm x}
\newcommand{\fxt}{\bm f(\bm x,t)}
\newcommand{\barw}{\Bar{\bm w}}
\subsection{General Form}
Let's consider the general diffusion process defined by SDEs in the following form (see more details in Appendix A and D.1 of \citet{DBLP:conf/iclr/0011SKKEP21}):
\begin{equation}
\label{eq:general_forward_sde}
    \text{d}\bm{x} = \bm f(\bm x,t)\text{d}t +\bm{G}(\bm x, t)d\bm w,
\end{equation}
where $\bm f(\cdot,t):\mathbb{R}^d\rightarrow\mathbb{R}^d$ and $\bm{G}(\cdot,t):\mathbb{R}^d\rightarrow\mathbb{R}^{d\times d}$. The corresponding reverse-time SDE is derived by~\citet{anderson1982reverse}:
\begin{equation}
\label{eq:general_reverse_sde}
    \td\bmx=\left\{ \fxt - \nabla_{\bmx} \cdot[\gxt\gxt^{\text T}] - \gxt\gxt^{\text T}\nabla_{\bmx} \log p_t(\bmx) \right\}\td t+\gxt\td\barw,
\end{equation}
where we refer $\nabla_{\bmx}\cdot \bm F(\bmx) := [\nabla_{\bmx}\cdot \bm f^1(\bmx),...,\nabla_{\bmx}\cdot \bm f^d(\bmx)]^{\text T}$ for a matrix-valued function $\bm F(\bm x):=[\bm f^1(\bmx),...,\bm f^d(\bmx)]^{\text T}$, and $\nabla_{\bmx}\cdot \bm f^i(\bmx)$ is the Jacobian matrix of $f^i(\bmx)$.
Then the ODE corresponding to Eq.~\ref{eq:general_forward_sde} has the following form:
\begin{equation}
\label{eq:general_ode}
    \td\bmx=\left\{ \fxt - \frac{1}{2}\nabla_{\bmx}\cdot[\gxt\gxt^{\text T}] -\frac{1}{2}\gxt\gxt^{\text T}\nabla_{\bmx}\log p_t(\bmx) \right\}\td t.
\end{equation}
\subsection{Derivation of Our ODE}
In this work, we adopt the Variance Preserving (VP) SDE~\cite{DBLP:conf/iclr/0011SKKEP21} to define the forward diffusion process:
\begin{equation}
    \text{d} \bm x = -\frac{1}{2}\beta(t)\bm x \text{d}t+\sqrt{\beta(t)}\text{d} \bm w,
\end{equation}
where the coefficient functions of Eq.~\ref{eq:general_forward_sde} are $\fxt=-\frac{1}{2}\beta(t)\bmx\in\mathbb{R}^d$ and $\gxt=\bm G(t)=\sqrt{\beta(t)}\bm I_d\in\mathbb{R}^{d\times d}$, independent of $\bm x$. Following Eq.~\ref{eq:general_reverse_sde}, the corresponding reverse-time SDE is derived as:
\begin{equation}
\begin{split}
    \td\bmx &= \left[ -\frac{1}{2}\beta(t)\bmx - \beta(t)\nabla_{\bmx}\cdot\bm I_d - \beta(t)\bm I_d\nabla_{\bmx}\log p_t(\bm x)  \right]\td t + \sqrt{\beta(t)}\bm I_d\td \barw\\
    & = \left[ -\frac{1}{2}\beta(t)\bmx - \beta(t)\nabla_{\bmx}\log p_t(\bm x)  \right]\td t + \sqrt{\beta(t)}\td \barw\\
    &= -\frac{1}{2}\beta(t)\left[\bmx + 2\nabla_{\bmx}\log p_t(\bm x) \right]\td t + \sqrt{\beta(t)}\td \barw,
\end{split}
\end{equation}
which infers to the Eq.~\ref{eq:sde}. Then, we derive the deterministic process (ODE) on the basis of Eq.~\ref{eq:general_ode}:
\begin{equation}
    \begin{split}
        \td\bmx &= \left[ -\frac{1}{2}\beta(t)\bmx -\frac{1}{2}\beta(t)\nabla_{\bmx}\cdot\bm I_d-\frac{1}{2}\beta(t)\bm I_d\nabla_{\bmx}\log p_t(\bmx) \right]\td t\\
        &=\left[ -\frac{1}{2}\beta(t)\bmx -\frac{1}{2}\beta(t)\nabla_{\bmx}\log p_t(\bmx) \right]\td t\\
        &=-\frac{1}{2}\beta(t)\left[\bmx +\nabla_{\bmx}\log p_t(\bmx) \right]\td t,
    \end{split}
\end{equation}
which gives the derivation of Eq.~\ref{eq:ode_x}.
\newpage
\section{Evaluation of Sample Selection Strategy}
\label{app:sample_selection}
As we stated in \S\ref{sec:implement}, we adopt a sample selection strategy for content-related generation tasks (text editing and generation with keywords).
Previous works also have similar strategies to improve the generation quality (i.e., PPLM~\cite{Dathathri2020Plug} and FUDGE~\cite{DBLP:journals/corr/abs-2104-05218}).

Since our latent model is trained by VAE objective, a sample $\bm x \in\mathcal{X}$ corresponds to a distribution $\mathcal{N}(\bm\mu,\bm\sigma^2)$ in $\mathcal{Z}$. Thus, we can search for better output by expanding the search space through sampling $\bm z_n\sim\mathcal{N}(\bm\mu,\bm\sigma^2)$, where $n=1,..., N$, and pick the best. 
Specifically, from ODE sampling, $\bm z(0)$ acts as the mean, and the variance $\bm \sigma^2$ is predefined. We generate $\bm z_n$ by sampling $\bm\epsilon_n$ from standard Gaussian:
\begin{equation}
    \bm z_n = \bm z(0) + \bm\sigma\odot\bm\epsilon_n, \quad \bm \epsilon_n\sim\mathcal N(\bm 0, \bm I).
\end{equation}

We decode each $\bm z_n$ and pick the best one according to the criterion of the task. We prefer the output that conforms to the desired attribute and achieves a high BLEU score with the source text for the text editing task. We want the output that contains the desired keyword or its variants for the generation with keywords. 

In our experiments (text editing and generation with keywords), we set $N=20$ as the default. To better demonstrate the strategy's improvement, we provide the quantitative and qualitative results towards different $N$.

\begin{figure}[ht]
\begin{minipage}[ht]{0.5\textwidth}
\centering
\small
\begin{tabular}{cccccc}
\toprule

\multirow{2}{*}{$N$}&Accuracy$\uparrow$&\multicolumn{3}{c}{Content$\uparrow$}&Fluency$\downarrow$\\\cmidrule(r){2-2}\cmidrule(r){3-5}\cmidrule(r){6-6}
& Sentiment & iBL & rBL & CTC & PPL\\\midrule
2&0.75&51.1&21.4&0.4737&26.3\\
4&0.82&50.6&22.0&0.4729&26.7\\
6&0.89&49.6&22.3&0.4729&26.2\\
8&0.9&50.5&22.2&0.4732&25.9\\
10&0.92&50.8&23.1&0.4730&26.2\\
12&0.93&51.4&23.2&0.4733&26.1\\
14&\underline{0.94}&51.4&23.0&0.4732&26.9\\
16&\underline{0.94}&52.4&23.4&0.4737&\underline{25.9}\\
18&\textbf{0.95}&\underline{52.6}&\underline{23.6}&\underline{0.4739}&\textbf{25.8}\\
20&\textbf{0.95}&\textbf{54.0}&2\textbf{4.2}&\textbf{0.4743}&\underline{25.9}\\
\bottomrule
\end{tabular}
  \captionof{table}{Automatic evaluation results towards to different $N$ on Yelp review dataset. 
  We mark the best \textbf{bold} and the second best \underline{underline}.}
  \label{tab:sample_selection}
\end{minipage}
\hfill
\begin{minipage}[ht]{0.48\textwidth}
\centering
\includegraphics[width=0.9\textwidth,page=1]{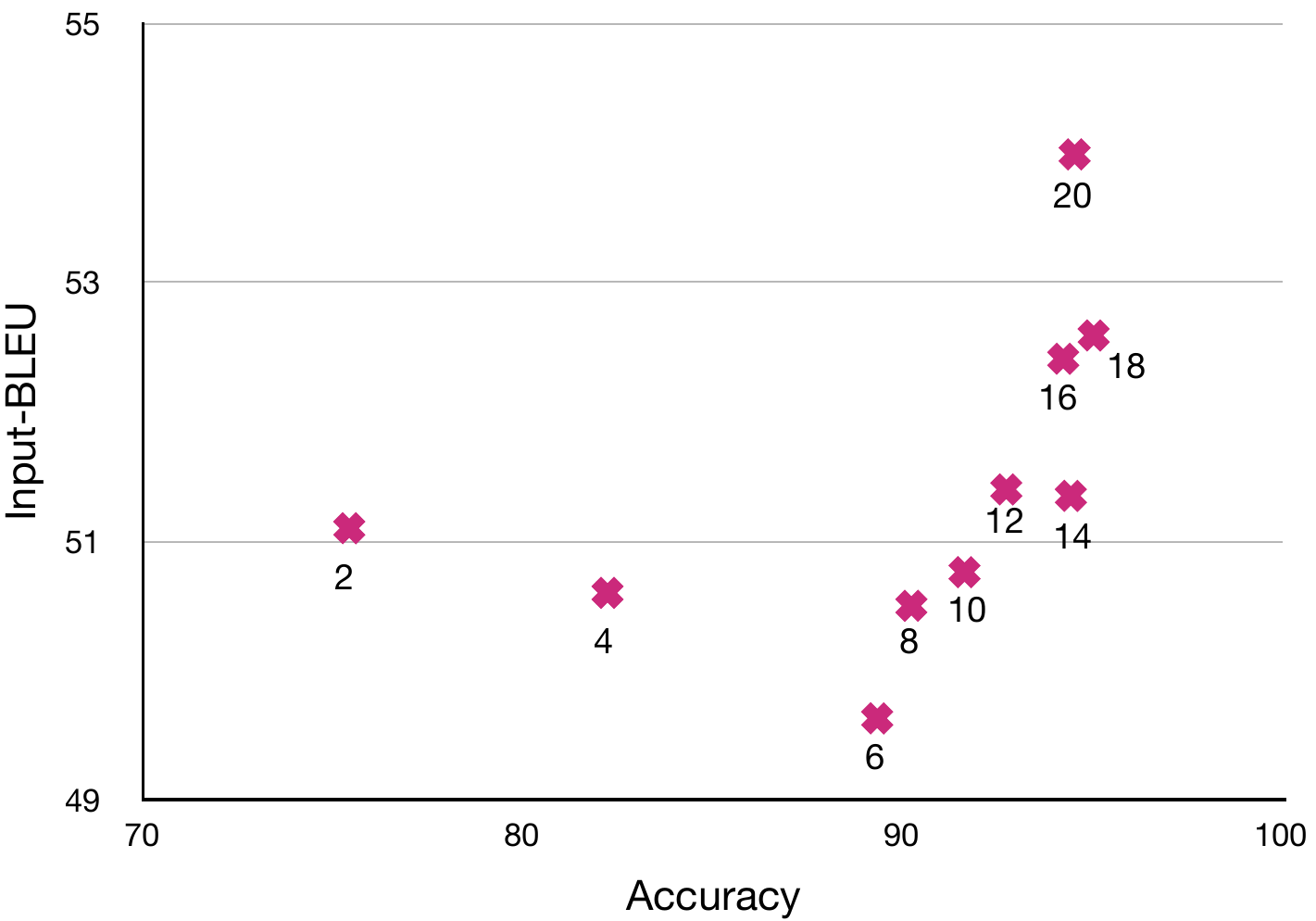}
\captionof{figure}{The trend of change of accuracy and input-BLEU as $N$ increases. The digit below each data point represents the corresponding $N$.}
\label{fig:sample_selection}
\end{minipage}
\end{figure}

We follow the same setting of text editing with single attribute on Yelp (\S\ref{app:example_text_edit_single}). The automatic evaluation results are shown in Table~\ref{tab:sample_selection}. As $N$ increases, all the metrics get improved. To reflect the trend of change in accuracy and content preservation, we plot Figure~\ref{fig:sample_selection}, which indicates that large $N$ gives better accuracy and better input-BLEU.

We also provide some examples in Table~\ref{tab:exmaple_sample_selection}. One observation is that all the outputs from the same source sequence describe similar scenarios but slightly differ in expression.
Thus, we can select the most suitable expression based on predefined rules. 

\begin{table}[ht]
    \centering
    \footnotesize
    \vspace{-19pt}
    \begin{tabular}{l|l}
    \toprule
Source&there is definitely not enough room in that part of the venue .\\
Target&there is so much room in that part of the venue\\\midrule
&there is definitely plenty of room in that perfect location .\\
&there is definitely no room enough in that venue to be the best part .\\
&there is definitely plenty of room right in that venue .\\
&there is definitely plenty of room right in the venue that needs .\\
&there is definitely plenty of room right in that venue .\\
&there is definitely enough room that can be right in the venue .\\
&there is definitely nothing better in room for that type of venue .\\
&there is definitely plenty of room in the right venue for that level .\\
&there is definitely nothing better in that room style of place .\\
&there is definitely a good room inside that best of all need in space .\\
&there is definitely plenty of room in the right level that is appropriate .\\
&there is definitely enough room in that right part of the venue .\\
&there is definitely plenty of room right in the deck that is needed .\\
&there is definitely enough room in that good atmosphere .\\
&there is definitely plenty of room in the right area , which is comfortable .\\
&there is definitely plenty of room in that perfect state of the place .\\
&there is definitely plenty of room that ideal in the location .\\
&there is definitely enough room in that perfect venue to all .\\
&there is definitely plenty of room in the right venue as well .\\
&there is definitely plenty of room available in the overall venue , too .\\
\midrule[1pt]
Source&it is n't terrible , but it is n't very good either .\\
Target&it is n't perfect , but it is very good .\\\midrule
&it is n't terrible , but it is very good also !\\
&it is very good , but it does n't even look great !\\
&it is n't terrible , but it is very good and definitely is good !\\
&it is n't great , but it is definitely very good !\\
&it is n't terrible , it is good and the menu is definitely great !\\
&it is n't terrible , but it is n't very good either .\\
&it is n't terrible , but it is very good also .\\
&it is n't terrible , but it is very good also !\\
&it is n't terrible , but it is definitely very good !\\
&it is very good , and it is n't terrible either .\\
&it is n't terrible , but it is very good and well made !\\
&it is very good , but it 's not really great either .\\
&it is n't terrible , but it is very good and well worth it .\\
&it is n't terrible , but it is definitely very good and good !\\
&it is n't terrible , but it is very good also !\\
&it is n't terrible , but it is very good and definitely is great !\\
&it is n't terrible , but it is very good also .\\
&it is n't terrible , but it is n't very good either .\\
&it is n't terrible , but it is very good also .\\
&it is n't terrible , but it is very good and always great !\\
\midrule[1pt]
Source&the food was pretty bad , i would not go there again .\\
Target&the food was great, i would go there again.\\\midrule
&he food was pretty good , i would go there again .\\
&the food was pretty good , i would def go there again !\\
&the food was pretty good , i would go again !\\
&the food was pretty good , i would go there again !\\
&the food was pretty good , i would definitely go there again .\\
&the food was pretty good , i would go back there again .\\
&the food was pretty good , i would definitely go back again .\\
&the food was pretty good , i would definitely go there again !\\
&the food was pretty good , i would definitely go there again .\\
&the food was pretty good , i would always go there again .\\
&the food was pretty good , i would go there again .\\
&the food was pretty good , i would not go there again .\\
&the food was pretty good , i would go there again .\\
&the food was pretty good , i would go back there again .\\
&the food was pretty good , i would go there again .\\
&the food was pretty good , i would definitely go there again !\\
&the food was pretty good , i would not go there again .\\
&the food was pretty good , i would definitely go there again .\\
&the food was pretty good , i would definitely go back again .\\
&the food was pretty good , i would go here again .\\
\bottomrule
    \end{tabular}
    \vspace{-7pt}
    \caption{Examples of sample selection strategy ($N=20$). 
    }
    \label{tab:exmaple_sample_selection}
\end{table}
\clearpage
\section{Distinguishing from Other Works}
\label{app:lace}
In this section, we outline the foundational and methodological differences that set \textsc{LatentOps} apart from models like LACE~\citep{nie2021controllable}.
The underlying motivation for our approach is fundamentally different. Text, unlike images, is characterized by discrete values and varying lengths, making it inherently more challenging to model. Given these complexities, there are only a handful of works exploring text operations within a condensed latent space. If we can fully comprehend this text latent space, we can align various textual tasks, such as generation and text editing, with operations in the latent domain.

LACE~\citep{nie2021controllable}, for instance, builds its foundation on pre-trained GANs. In order to train their classifiers, class labels for latent vectors are essential. This necessitates the use of external classifiers to retrieve the class labels of the latent vectors, with human intervention required to filter out subpar samples. Our approach, on the other hand, effectively adapts large pre-trained LMs to the VAE framework. For specific datasets, the VAE is not only efficient but also expeditious in its training. Thanks to the bi-directional mapping between the latent space and text space, we can directly train the classifier using the marginal distribution.

Furthermore, LACE establishes a joint distribution in the image space and then transitions to the latent space using the reparameterization trick. Contrarily, our model defines its joint distribution directly within the text latent space.

Additionally, the architecture of LACE, built upon the GAN latent space, necessitates certain specialized regularization terms in its energy function for different tasks to optimize performance. Our model benefits from the more structured latent space provided by the VAE, allowing our energy function to remain straightforward and consistently aligned with our practical definitions.

\section{More Details and Results of Experiments}
In this section, we provide more details and results of the experiments (\S\ref{sec:exp}).
\subsection{Setup}
\label{app:setup}
The Yelp dataset and Amazon dataset contain 443K/4K/1K and 555K/2K/1K sentences as train/dev/test sets, respectively.
Since Yelp and Amazon datasets\footnote{\url{https://github.com/lijuncen/Sentiment-and-Style-Transfer}}\footnote{The datasets are distributed under CC BY-SA 4.0 license.} are mainly developed for sentiment usage, we annotate them with a POS tagger to get the tense attribute
to test the ability of our model that can be extended to an arbitrary number of attributes. 
Besides, we also use GYAFC dataset~\cite{DBLP:conf/naacl/RaoT18} to include the formality attribute. Note that the GYAFC dataset
has somewhat different domains from Yelp/Amazon, which can be used to test our model's out-of-domain generalization ability. All the datasets are in English.

We adopt BERT-small\footnote{The BERT model follows the Apache 2.0 License.} and GPT2-large\footnote{The GPT2 model follows the MIT License.}  as the encoder and decoder of our latent model, respectively. The training paradigm follows \S\ref{sec:implement}, and some training tricks~\cite{li-etal-2020-optimus} (i.e., cyclical schedule for KL weight and KL thresholding scheme) are applied to stabilize the training of the latent model. All the attributes are listed in Table~\ref{tab:operators}. 
All the models are trained and tested on a single Tesla V100 DGXS with 32 GB memory.
Input-BLEU, reference-BLEU and self-BLEU are implemented by nltk~\cite{DBLP:books/daglib/0022921} package. 


We employ a BERT classifier to determine attribute accuracy, serving as a metric for the evaluation of the success rate. More precisely, we finetune BERT-base models dedicated to classification tasks using the respective dataset. For instance, when evaluating sentiment, the classifier tailored for the Yelp dataset registers accuracies of 97.1\% and 97.3\% on the dev and test sets, respectively. Meanwhile, for the Amazon dataset, the sentiment classifier records accuracies of 86.9\% and 85.7\% on the dev and test sets.

In our experiments of generation with single attribute, we also incorporate the MAUVE metric~\citep{pillutla2021mauve}, an automatic measure of the gap between neural text and human text for text generation. In alignment with the official recommendations associated with MAUVE, we select a random subset of 10,000 sentences from the training set to serve as reference sentences.

For the operator (classifier) $f_i(\bm z)$, we adopt a four-layer MLP as the network architecture as shown in Table~\ref{tab:arc_classifier}. Since the number of trainable parameters of the classifier is small, it is rapid to train and sample. 
\begin{table}[ht]
    \centering
    \begin{tabular}{l|l |l}
    \toprule
        Style & Attributes  & Dataset\\\midrule
         Sentiment& Positive / Negative  &Yelp, Amazon\\
         Tense& Future / Present / Past & Yelp \\
         Keywords& Existence / No Existence & Yelp\\
         Formality & Formal / Informal & GYAFC \\
         \bottomrule
    \end{tabular}
    \caption{All attributes and the corresponding dataset are used in our experiments.} 
    \label{tab:operators}
\end{table}

\begin{table}[ht]
    \centering
    \small
    \begin{tabular}{lllll}
        \toprule
        Input & Layer 1 & Layer 2 & Layer 3 & Layer 4\\\midrule
         $\bm z\in\mathbb{R}^{64}$&
         Linear 43, LeakyReLU&
         Linear 22, LeakyReLU&
         Linear 2, LeakyReLU&
         Linear \#logits\\\bottomrule
    \end{tabular}
    \caption{The architecture of the attribute classifier.}
    \label{tab:arc_classifier}
\end{table}
\paragraph{Observations on Scalability}
In our experiments with different encoder and decoder scales, several observations emerged. Firstly, while we evaluated various encoder models, including BERT, RoBERTa, and other pre-trained language models (PLMs) of different scales, the distinctions in performance were minimal. In this context, BERT-small proved sufficiently robust, serving as an effective encoder for the VAE framework. Secondly, the scale of the decoder was observed to significantly influence performance. We examined a spectrum of models, ranging from GPT2-base to GPT2-xl. Through these tests, GPT2-base and GPT2-large emerged as the optimal choices, providing a harmonious blend of performance results and computational efficiency.
\paragraph{Stability of VAE training}
The stability of VAE training has benefited from recent innovations, as evidenced by works like \citet{li-etal-2020-optimus}. Our empirical observations, corroborated by subsequent research such as \citet{hu2021causal}, attest to these advancements. In addition, our approach's constrained parameter training further enhances this stability. As a result, training the VAE has become less of a challenge or bottleneck than before.
\subsection{Generation with Compositional Attributes}
The section is a supplement of \S\ref{sec:cg_exp}, we give more details of experimental configuration, generated examples and discussion.
\subsubsection{More Details of Baselines}
\label{app:generation_baselines}
We compare our method with PPLM~\cite{Dathathri2020Plug}, FUDGE~\cite{DBLP:journals/corr/abs-2104-05218}, and a finetuned GPT2-large~\cite{gpt2}. PPLM and FUDGE are plug-and-play controllable generation approaches on top of an autoregressive LM as the base model. For fair comparison (\S\ref{sec:vae_training}), we obtain the base model by finetuning the embedding layer and the first transformer layer of pretrained GPT2-large on the Yelp review dataset with unlabeled data. All the classifiers/discriminators of PPLM, FUDGE and our \textsc{LatentOps} are trained by a small subset of the original dataset (200 labeled data instances per class). 
\paragraph{PPLM}  requires a discriminator attribute model (or bag-of-words attribute models) learned from a pretrained LM's top-level hidden layer. At decoding, PPLM modifies the states toward the increasing probability of the desired attribute via gradient ascent. We only consider the discriminator attribute model, which is consistent with other baselines and ours. We follow the default setting of PPLM, and for each attribute, we train a single layer MLP as the discriminator.  
\paragraph{FUDGE} has a discriminator that takes in a prefix sequence and predicts whether the generated sequence would meet the conditions. FUDGE could control text generation by directly modifying the probabilities of the pretrained LM by the discriminator output. 
We follow the architecture of FUDGE and train a discriminator for each attribute.
Furthermore, we tune the $\lambda$ parameter of FUDGE which is a weight that controls how much the probabilities of the pretrained LM are adjusted by the discriminator, and we find $\lambda$=10 yields the best results. We follow the default setting of FUDGE, and for each attribute, we train a three-layer LSTM followed by a Linear as the discriminator.
\paragraph{GPT2-FT} is a finetuned GPT2-large model that is a conditional language model, not plug-and-play. Specifically, we train an external classifier for the out-of-domain attribute (i.e., formality) to annotate all the data in Yelp. For tense, we use POS tagging to annotate the data automatically. Then we finetune the embedding layer and the first layer of GPT2-large by the labeled data. Since GPT2-FT is fully-supervised and not plug-and-play, it is not comparable with other baselines and ours, and we only use it for reference.
\subsubsection{More Discussion of Generation with Compositional Attributes}
\label{app:generation_compositional}
\paragraph{Discussion of Quantitative Results}
As we state in \S\ref{sec:generation_composable}, our method is superior to baselines. We want to discuss the results in Table~\ref{tab:multi_cg}.

For success rate, our method dramatically outperforms FUDGE and PPLM as expected since both control the text by modifying the outputs (hidden states and probabilities) of PLM, which includes the token-level feature and lacks the sentence-level semantic feature. On the contrary, our method controls the attributes by operating the sentence-level latent vector, which is more suitable. 

For diversity, since our method bilaterally connects the discrete data space with continuous latent space, which is more flexible to sample, ours gains obvious superiority in diversity. Conversely, PLMs like GPT2, which is the basis of PPLM and FUDGE, are naturally short of the ability to generate diverse texts. They generate diverse texts by adopting other decoding methods (like top-k), which results in the low diversity of the baselines. 

For fluency, we calculate the perplexity given by a finetuned GPT2, which processes the same architecture and training data of PPLM and FUDGE, so naturally, they can achieve better perplexity even compared to the perplexity of test data and human-annotated data. Moreover, our method only requires an Extra Adapter to guide the fixed GPT2, and our fluency is in a regular interval, a little higher than the perplexity of human-annotated data.

Since GPT2-FT is trained with full joint labels (all the data has all three attribute labels), it can achieve a reasonable success rate, and ours is comparable. Moreover, consistent with PPLM and FUDGE, GPT2-FT can achieve good perplexity but poor diversity due to the sampling method.

\paragraph{Discussion of Qualitative Results}
We provide some generated examples in Table~\ref{tab:cg_examples} to raise a more direct comparison. Consistent with the quantitative results, it is difficult for FUDGE to control all the desired attributes successfully, although GPT2-FT and ours perform well. For diversity, it is evident that FUDGE and GPT2-FT prefer to generate short sentences containing very little information. Some words appear highly, yet ours gives a more diverse description. Regarding fluency, since FUDGE and GPT2-FT tend to generate simple sentences, they can obtain better perplexity readily. However, ours is inclined to generate more informative sentences. In conclusion, there is a trade-off between diversity and fluency. It can be handled well by ours, but for the baselines, they pursue fluency too much and lose diversity.

\begin{table}[ht]
\centering
\scriptsize
\vspace{-15pt}
 \setlength\tabcolsep{1.5pt}
\begin{tabular}{m{0.4\textwidth}m{0.4\textwidth}}
\toprule
\textbf{Positive + Present + Formal}&\textbf{Negative + Past + Inormal}  \\\midrule
 \multicolumn{1}{l}{GPT2-FT:}&\multicolumn{1}{l}{GPT2-FT:}\\
    \quad the staff is friendly and helpful.&\quad didn't bother with the food and just walked out.\\
    \quad i love it \senc{here}. \senc{[Informal]}&\quad just not a good place for me. \senc{[No Tense]}\\
    \quad this {is} the place to go for {traditional} chinese food.&\quad not a fan of this place. \senc{[No Tense]}\\
    \quad highly recommend them. \senc{[Informal]}&\quad just not good. \senc{[No Tense]}\\
    \quad the menu {is} small but {very nice}.&\quad horrible! \senc{[No Tense]}\\
    \quad it{'s} a {great} place.&\quad oh and the cake was way too salty.\\
    \quad i {highly {recommend}} this place.&\quad but we didn't even finish it.\\
    \midrule
 \multicolumn{1}{l}{PPLM:}&\multicolumn{1}{l}{PPLM:}\\
    \quad i love this store and the service is always friendly and courteous.&\quad i ordered delivery... what?\\
    \quad the staff was so friendly \senc{\&} helpful!\senc{[Informal]}&\quad \senc{great} service. \senc{[No Tense]}\\
    \quad the place is clean.&\quad this place was terrible!\\
    \quad the best french bakery i have ever been to in las vegas!&\quad the service was horrible horrible horrible!\\
    \quad this place \senc{was} a gem!&\quad i ordered the ribs and brisket tacos and it was very bland. \senc{[Formal]}\\
    \quad she does love to make suggestions and i appreciate that.&   \quad the staff was very apologetic \senc{and apologetic} and \senc{refund} my \$ \_num\_ for the oil change \senc{[Formal]}\\
     \quad they also always remember us \senc{and always always} get us right in and always have good prices.&\quad i ordered pizza and wings from brooklyn's and they were all out of ranch. \senc{[Formal]}\\
    \midrule
 \multicolumn{1}{l}{FUDGE:}&\multicolumn{1}{l}{FUDGE:}\\
    \quad great for breakfast or a nice lunch. \senc{[Informal]}& \quad came to phoenix from new jersey last weekend...!\\
    \quad great location. \senc{[Informal]}& \quad food was ok, but service was terrible!\\
    \quad their staff is friendly, professional, and the facility is clean and comfortable. & \quad usually the service was \senc{good} and the food was \senc{good no complaints}.\\
    \quad great. \senc{[Informal \& No Tense]}& \quad food was ok but our waiter was awful.\\
    \quad great place for lunch or a date. \senc{[No Tense]}& \quad {c} was \senc{amazing}.\\
    \quad great place! \senc{[Informal \& No Tense]}& \quad {c} was \senc{so good} and i \senc{highly recommend}. \\
    \quad great food. \senc{[Informal \& No Tense]}& \quad {ch} was the only reason i stayed for the night.\\
    \midrule
 Ours:&Ours:\\
    \quad the food is clearly great , as they are always tasty .&\quad everything was a bit cold but anyways , i ordered them !\\
    \quad they are really knowledgeable , what draws me .&\quad anyway i had the worse experience !\\
    \quad the shop is authentic , their hair is great .&\quad looked like i was n't even paid this money !\\
    \quad the food is always unique with well spiced .&\quad ( had no job in \textit{\_num\_} months from cali . )\\
    \quad that is a great form of customer service .&\quad i waited at the room \& got \textit{\_num\_} people yelling ?\\
    \quad they have very professional people who are worth their service .&\quad ( i didnt get this at all times )\\
    \quad i love living there as does my clients .&\quad they had me cold a lot !\\
    \midrule[1pt]
\textbf{Negative + Future + Formal}&\textbf{Positive + Past + Informal}\\
\midrule
GPT2-FT:& GPT2-FT: \\
	\quad i will not be back.&\quad good prices too! \senc{[No Tense]}\\
	\quad would not recommend this location to anyone.&\quad i even liked the cheese curds....\\
	\quad would not recommend them for any jewelry or service.&\quad hands down the best sushi i've had in a while.\\
	\quad if i could give this place zero stars, i would.&\quad just a great shop! \senc{[No Tense]}\\
	\quad if i could give no stars, i would.&\quad my friend had a good time.\\
	\quad i would not recommend this place to anyone.&\quad got ta love that!\\
	\quad i {can} not get my medication on time.&\quad really good service, super fast and friendly. \senc{[No Tense]}\\
	\midrule

PPLM:&PPLM:\\
	\quad i could not recommend them at all.&\quad i ordered a great deal at a very good sushi restaurant tonight. \senc{[Formal]}\\
	\quad i \senc{could not} believe this was \senc{not good}!&\quad it \senc{is} light and airy and \senc{has} very few after tastes of smoke or heat.\\
	\quad this \senc{was a big deal}, because the food \senc{was great}. &\quad i loved it so much i had to get the other salad!\\
	\quad i could not recommend them.&\quad the staff at my table had the best service ever!\\
	\quad i will not be back.&\quad we've had some really great ones too.\\
	\quad the food \senc{was} mediocre.&\quad i \senc{love} everything and \senc{would} highly recommend!\\
	\quad they \senc{were} not.&\quad they did a fabulous job of putting me on a diet for the first time in my life! \senc{[Formal]}\\
	\midrule

FUDGE:&FUDGE:\\
    \quad not a great pizza to get a great pie! \senc{[No Tense]}& \quad thanks was definitely great!\\
    \quad however, this place \senc{is pretty good}. & \quad went and spent the whole night here and had a blast!\\
    \quad i \senc{have never} seen anything like these. & \quad she loved the food and service!\\
    \quad will definitely return. & \quad went and the food was good, \senc{nothing special}.\\
    \quad i would have loved to have a \senc{nice} lunch here. & \quad he was friendly, knowledgeable and very helpful! \\
    \quad they \senc{don't have} any of the ingredients they should. & \quad great beer was amazing!\\
    \quad \senc{do} not go here for the food. & \quad went on to eat and was \senc{very disappointed} with our food!\\
	\midrule

Ours:&Ours:\\
	\quad i would not believe them to stay .&\quad everything was \senc{hot} and incredibly good !\\
	\quad i will never be back .&\quad plus they had a great and fresh meal here !\\
	\quad i would not recommend her to anyone in the network .&\quad fresh mozzarella was great in general !\\
	\quad they will not think to contact me for any reason .&\quad the veggies and omelette were great !\\
	\quad i should not risk coming to this establishment .&\quad great service and enjoyed our out day meal\\
	\quad i would not waste more time in henderson .&\quad i ended up getting a great meal ( i loved it ! )\\
	\quad i doubt i would 've ever been to this airline .&\quad ( she got a job for me !   \\
	\bottomrule
    \end{tabular}
    \caption{More examples of generation with compositional attributes. We mark failed spans in \senc{red}.}
    \label{tab:cg_examples}
\end{table}

\clearpage
\subsubsection{Results of Generation with Compositional Attributes and Keywords}
\label{app:generation_keyword}

We regard keywords as an attribute of the text sequence. To prepare the data, we extract all verbs, nouns, and variants that appeared in the Yelp review dataset, filter out the sentiment-related words\footnote{\url{http://www.cs.uic.edu/~liub/FBS/opinion-lexicon-English.rar}}, and construct the training data. 
Then, we obtain 613 keywords listed in Table~\ref{tab:all_keywords}.
We treat each keyword (e.g., \textit{have}) and their variants (e.g., \textit{had} or \textit{has}) equally without discrimination. 
Moreover, for each keyword, we randomly select 220 sentences where the keyword exists and 220 sentences that do not include the keyword as the training data (200) and test data (20).
Since we have 3,678 combinations of keyword, sentiment and tense, we adopt a pretrained GPT2 base model as the decoder to accelerate the process.

%
We conduct the experiments of single keyword and keyword combining with other attributes (sentiment and tense). 
We first give the automatic evaluation results in Table~\ref{tab:lexical_cg}. We list the average results of each combination of keywords, sentiment and tense. All success rates, diversity and fluency, are at a high level. To make the results more intuitive, we also give some generated examples in Table~\ref{tab:lexical_1}.
\begin{table}[ht]
\small
\centering
\vspace{-8pt}
\begin{tabular}{lcccccc}
\toprule
\multirow{2}{*}{Attributes}& \multicolumn{4}{c}{Accuracy$\uparrow$} &  Fluency$\downarrow$ & Diversity$\downarrow$\\
\cmidrule(r){2-5}\cmidrule(r){6-6}\cmidrule(r){7-7}
&Keyword & Sentiment & Tense & G-Mean &PPL & sBL    \\
\midrule
{Keyword}  &{0.98} &-&-&{0.98} & 21.7 & {10.6} \\
                 \cmidrule{2-7}
{+ Sentiment}   &{0.94}& {0.96}& -& {0.95}& 21.3& {10.8}\\
                 \cmidrule{2-7}
{\ \ + Tense}   &{0.93}& {0.9}& {0.93}  &{0.92}   & 19.7            & 10.9\\
                 \bottomrule
\end{tabular}
\vspace{-5pt}
\caption{Results of generation with compositional attributes and keywords. 
}
\label{tab:lexical_cg}
\vspace{-10pt}
\end{table}

\begin{figure}[ht]
\vspace{-5pt}
\begin{minipage}[ht]{.51\linewidth}
\centering
\scriptsize
\begin{tabular}{l}
\toprule
\textbf{Keyword}: \textit{expectation}\\
the prices were excellent and exceeded our \tenc{expectations} .
\\
five stars , affordable and reasonable pricing exceeded my \tenc{expectations} .
\\
i 've had four peaks meal from my \tenc{expectations} and i have not disappointed .
\\
you are crazy close to my \tenc{expectations} !
\\
the flavors have never been above \& beyond \tenc{expectations} .\\\midrule
\textbf{Keyword}: \textit{expectation} + \textbf{Sentiment}: Negative\\
the appetizers were \tenc{completely lower expectations} .
\\
i would give this restaurant \textit{\_num\_} \tenc{zero expectations} in terms of our entrees .
\\
it \tenc{was n't that impressive} and \textit{\_num\_} \tenc{declined my expectations} .
\\
there were \tenc{zero expectations} .
\\
but my \tenc{expectations} were \tenc{lower than zero stars} .
\\\midrule
 \textbf{Keyword}: \textit{expectation} + \textbf{Sentiment}: Negative + \textbf{Tense}: Past\\
there \tenc{were} so \tenc{low expectations} throughout the end .
\\
the food \tenc{was} ok , but my \tenc{expectations were high} to top notch .
\\
during the event we \tenc{were already disappointed} with the  \tenc{expectations} .
\\
we  \tenc{arrived} \textit{\_num\_} months ago and my  \tenc{expectation was overcharged} .
\\
again , the initial estimate of course \tenc{had not gotten my expectations} and declined .
\\\midrule
 \textbf{Keyword}: \textit{expectation} + \textbf{Sentiment}: Negative + \textbf{Tense}: Present\\
the prices  \tenc{are} really low and restaurants  \tenc{are not above expectations} .
\\
there  \tenc{is} almost  \tenc{no flavor} in my  \tenc{expectations} .
\\
the chips and salsa  \tenc{are far below their expectations} and  \tenc{lack of manners} .
\\
it  \tenc{'s} about the  \tenc{expectations lower than zero} .
\\
the food in american restaurants  \tenc{do not exceed your expectations} .
\\\midrule
 \textbf{Keyword}: \textit{expectation} + \textbf{Sentiment}: Negative + \textbf{Tense}: Future\\
i  \tenc{would not come back} to any  \tenc{expectations} of this restaurant .
\\
it \tenc{would n't be exceeded my expectations} at any point .
\\
i \tenc{would n't want you to have any expectations} in this hotel .
\\
honestly i \tenc{would n't have lower expectations} before .
\\
i \tenc{would not expect superior} from my \tenc{expectation} .\\
\bottomrule
\end{tabular}
\end{minipage}
\hfill
\begin{minipage}[ht]{.47\linewidth}
\centering
\scriptsize
\begin{tabular}{|l}
\toprule
\textbf{Keyword}: \textit{accommodate}\\
    staff was nice and \tenc{accommodating} a timely manner .\\
    he is always nice and \tenc{accommodating} .\\
    the service is wonderful and the facility is clean and \tenc{accommodating} .\\
    nicely crowded , along with a great \tenc{accommodating} staff !\\
    she is friendly and willing to \tenc{accommodate} any type of questions .\\\midrule
\textbf{Keyword}: \textit{accommodate} + \textbf{Sentiment}: Positive\\
    staff is very \tenc{nice} and the servers are \tenc{friendly and accommodating} .\\
    everyone was very \tenc{friendly and accommodating} with a ton of energy !\\
    tamara was extremely \tenc{nice and accommodating} .\\
    everyone seemed to talk with  \tenc{accommodating} .\\
    he made a  \tenc{wonderful massage} to  \tenc{accommodate} my kids .\\\midrule
 \textbf{Keyword}: \textit{accommodate} + \textbf{Sentiment}: Positive + \textbf{Tense}: Past\\
    they  \tenc{were} really  \tenc{nice} and  \tenc{made} to  \tenc{accommodate} me with a great energy .\\
    the everyone \tenc{was} very \tenc{nice} and the hospitality \tenc{was accommodating} as well !\\
    the whole family \tenc{was accommodating} and we \tenc{enjoyed} the round !\\
    the staff \tenc{was} always \tenc{friendly} and \tenc{accommodating} with \tenc{great suggestions} .\\
    thanks , the hostess \tenc{was extremely helpful} and \tenc{accommodating} .\\\midrule
 \textbf{Keyword}: \textit{accommodate} + \textbf{Sentiment}: Positive + \textbf{Tense}: Present\\
    they \tenc{are friendly} and \tenc{helpful} , and the pricing \tenc{is easy} to \tenc{accommodate} .\\
    the staff \tenc{is amazing} and very \tenc{accommodating} and the owners \tenc{are wonderful} .\\
    everyone i\tenc{s super nice} and \tenc{accommodating} !\\
    the servers \tenc{are always accommodating} and \tenc{helpful} !\\
    the venue \tenc{is quite accommodating} , and a \tenc{great happy atmosphere} .\\\midrule
 \textbf{Keyword}: \textit{accommodate} + \textbf{Sentiment}: Positive + \textbf{Tense}: Future\\
    they \tenc{will} definitely \tenc{stay close} to \tenc{accommodate} us !\\
    they \tenc{would} very \tenc{reasonable} to \tenc{accommodate} you in any condition !\\
    hopefully , they \tenc{will definitely be accommodated} with our family !\\
    they \tenc{would} be able to \tenc{accommodate} you at any location .\\
    i \tenc{would definitely recommend} this firm to \tenc{accommodate} us !\\
\bottomrule
\end{tabular}
\end{minipage}
    \vspace{-5pt}
    \captionof{table}{Examples of generation with compositional attributes with keywords (\textit{expectation} and \textit{accommodate}).We mark the spans that conform to desired attributes in \tenc{blue}. }
    \label{tab:lexical_1}
\end{figure}

\subsubsection{Results of Generation with Single Attribute}
\label{app:generation_single}
Table~\ref{tab:single-cg} gives the results of single-attribute conditional generation. Our method dramatically outperforms PPLM and FUDGE for all attributes on the accuracy,  exceeding 94\%. The diversity and fluency exhibited by our method align well with the results from multi-attribute evaluations. The MAUVE metric is designed to quantify the information divergence between the distribution inferred by the text generation model and the actual data distribution. Our results further underscore the efficacy of our approach, suggesting that the distribution learned by our method more closely approximates the real data distribution.
\begin{table}[ht]
\centering
\small
\vspace{-7pt}
\begin{tabular}{ccccccc}
\toprule
Attributes   & Methods & Accuracy$\uparrow$  & LogVar$\downarrow$ & Fluency (PPL)$\downarrow$ &Diversity (sBL) $\downarrow$  & MAUVE$\uparrow$ \\\midrule
\multirow{5}{*}{Sentiment}     &GPT2-FT& 0.98 & -11.31 & 10.6 & 23.8 &0.049 \\\cmidrule{2-7}
& PPLM & 0.86 & -4.68 & 11.8 & 31.0  & 0.102\\ 
& FUDGE & 0.77 & -2.97 & \textbf{10.3} & 27.2&0.050\\  
& Ours &\textbf{0.99} & \textbf{-Inf} & 30.4 & \textbf{13.0} &\textbf{0.807}\\\midrule 

\multirow{5}{*}{Tense}&GPT2-FT& {0.97} & {-9.33} & {10.0} & 31.0&0.057 \\
\cmidrule{2-7} 
& PPLM &0.6 & -3.30 & 13.9 & 27.8  &0.093\\ 
& FUDGE & 0.77 & -3.11 & \textbf{10.9} & 37.6 &0.085\\ 
& Ours& \textbf{0.96} & \textbf{-6.8} & 36.7&\textbf{9.5}  &\textbf{0.847}\\\midrule  

\multirow{5}{*}{Formality}   &GPT2-FT& 0.88 & -5.75 & 14.9 & 18.0 &0.080\\\cmidrule{2-7} 
& PPLM &0.62 & -2.43 & 14.8 & 24.8 &0.083  \\ 
& FUDGE& 0.59 & -2.16 & \textbf{11.2} & 28.6 &0.054 \\  
& Ours& \textbf{0.97} & \textbf{-7.82} & 36.3 & \textbf{12.0} &\textbf{0.774} \\\bottomrule  
\end{tabular}
\caption{Automatic evaluation results of generation with single attribute. We show the natural logarithm of variance (LogVar) of accuracy, since the original
scale is too small for demonstration.}
\label{tab:single-cg}
\vspace{-10pt}
\end{table}
\subsection{Text Editing}
The section is a supplement of \S\ref{sec:tst_exp}, we give more details of experimental configuration, generated examples and discussion.
\subsubsection{More Details of Baselines}
\label{app:baseline_text_edit}
For text editing, we experiment with three settings--sequential attribute editing, compositional attributes editing and single attribute editing. 

We compare with several recent state-of-the-art methods:
B-GST~\cite{DBLP:conf/emnlp/SudhakarUM19}, Style Transformer (STrans)~\cite{DBLP:conf/acl/DaiLQH19}, DiRR~\cite{DBLP:conf/naacl/LiuNW21}, Tag\&Gen (T\&G)~\cite{DBLP:conf/acl/MadaanSPPNYSBP20}, and fine-grained style transfer (FGST)~\cite{DBLP:conf/aaai/LiuFZPL20}. 
The outputs of baselines are obtained from their official repositories except for FUDGE. 
Since FUDGE relies on a PLM, we finetune a GPT2 as a reconstruction model as the base model.

FUDGE is the sole model that could handle compositional attributes. Therefore, we compare with FUDGE in the compositional attributes setting. Furthermore, we tune the $\lambda$ parameter of FUDGE which is a weight that controls how much the probabilities of the pretrained LM are adjusted by the discriminator, and we find $\lambda$=100 yields the best results. We compare with all baselines in the single attribute setting.
\subsubsection{Examples of Sequential Editing}
\label{app:example_seq_edit}
We provide more examples of the Sequential Editing (\S\ref{sec:sequential_edit}) experiment in Table~\ref{tab:seq_edit}, where the first two examples are the same as in \ref{tab:seq_edit_part}. Our method can sequentially edit the source text to desired attributes more smoothly and consistently.

In the first example, FUDGE fails on all three edits, Style Transformer introduces \textit{ate}, which leads to grammatical mistakes and loss of critical information (\textit{flowers)}. Our method can edit the source text step-by-step successfully. 

In the second example, FUDGE fails all edits again and introduces irrelevant information (\textit{thing's}). Furthermore, Style Transformer nearly fails in all edits. Our method could generate both fluent and content-relevant sentences. 

In the third example, we consider editing the source to formal, positive and past. FUDGE and Style Transformer only succeed in introducing the positive sentiment, and FUDGE also introduces some redundant information (\textit{to get away from the strip}). Ours first extends the source to be formal, then changes the sentiment (\textit{horrible} to \textit{amazing}) and tense (\textit{is} to \textit{was}), sequentially.

In the last example, FUDGE fails all edits. Although Style Transformer succeeds in sentiment transfer, the generated sentence is not grammatically correct. Ours could generate eligible and fluent sentences.

\begin{table}[ht]
    \centering
\small
    \begin{tabular}{ll}
    \toprule
          Source & the flowers and prices were great . \\
         \midrule
           FUDGE:&\\
         + informal&the flowers and prices were great. \senc{[Formal]}\\
         \quad + negative& \senc{garlic pizza} and prices were \senc{great}.\\
         \quad\quad + present&\senc{garlic pizza} and prices \senc{were great}.\\
          STans:&\\
         + informal&the flowers and prices were great ?\\
         \quad+ negative&the \senc{ate} and prices were terrible ?\\
         \quad\quad+ present&the \senc{ate} and prices are terrible ?\\
          Ours:& \\
         + informal& and the flowers and prices were great ! \\
         \quad+ negative& and the flowers and prices were terrible !\\
         \quad\quad+ present& and the flowers and prices are terrible !\\\midrule
         Source & best korean food on this side of town .\\\midrule
         FUDGE:&\\
         + informal&best korean food on this side of town. \senc{[Formal]}\\
         \quad+ negative&\senc{thing's best} korean food on this side of town.\\
         \quad\quad+ present &\senc{thing's best} korean food on this side of town. \senc{[No Tense]}\\
          STans:&\\
         + informal&best korean food on this side of town \senc{korean food} . \senc{[Formal]}\\
         \quad+ negative&only korean food on this side of town \senc{korean food} .\\
         \quad\quad+ present&only korean food on this side of town \senc{korean food} . \senc{[No Tense]}\\
          Ours:& \\
          + informal&best korean food on this side of town ! \\
         \quad + negative& worst korean food on this side of town ! \\
          \quad\quad+ present& this is worst korean food on this side of town !\\\midrule
         Source & horrible . \\\midrule
          FUDGE:& \\
          + formal&horrible! \senc{[Informal]}\\
          \quad+ positive&great place \senc{to get away from the strip}.\\
          \quad\quad+ past&great place \senc{to get away from the strip}. \senc{[No Tense]}\\
          STrans:& \\
          + formal&horrible . \senc{[Informal]}\\
          \quad+ positive&wonderful .\\
          \quad\quad+ past& wonderful .\senc{[No Tense]}\\
           Ours:&\\
          + formal&service is completely horrible . \\
          \quad+ positive&service is completely amazing .\\
          \quad\quad+ past& service was completely amazing .\\\midrule
         Source & it is a garbage , and nobody does really care !\\\midrule
         FUDGE:& \\
          + informal &it is a garbage , and nobody does really care ! \senc{[Formal]} \\
          \quad+ positive &it is \senc{always a garbage} , and \senc{nobody} does really \senc{care} !\\
          \quad\quad+ future &it \senc{is always a garbage} , and \senc{nobody does} really \senc{care} !\\
          STrans:& \\
          + informal & it is a garbage , and nobody does really care ! \senc{[Formal]}\\
          \quad+ positive &it is a smile , and {high} does really care !\\
          \quad\quad+ future & it \senc{is} a smile , and {high} \senc{does} really care !\\
           Ours: & \\
          + informal & ( it is garbage services ... no crap !\\
          \quad+ positive & ( the delivery service is excellent ! )\\
          \quad\quad+ future & it is the first delivery service i will get ! \\\bottomrule
    \end{tabular}
    \caption{Examples of sequential editing. We mark
failed spans in \senc{red}.}
    \label{tab:seq_edit}
\end{table}

\clearpage
\subsubsection{Examples of Text Editing with Compositional Attributes}
\label{app:example_text_edit_compositional}
We provide some examples of Text Editing with Compositional Attributes (\S\ref{sec:text_edit_compositional}) in Table\ref{tab:example_tst_yelp}.
\begin{table}[ht]
    \centering
    \footnotesize
    \begin{tabular}{lp{13cm}}
\toprule
Source&so basically tasted watered down .\\
Human&it didn't taste watered down at all.\\
\midrule
FUDGE&\senc{once every couple months, we get a new car - }so basically tasted \senc{watered down}.\\
\quad + Past&such \senc{basically} tasted watered down.\\
\quad + Present&such \senc{basically tasted} watered down.\\
\quad + Future&very \senc{watered down}.\\
\midrule
Ours&so basically tasted \tenc{delicious} .\\
\quad +  Past&so \tenc{nicely tasted} watered down .\\
\quad +  Present&so basically \tenc{tastes delicious} .\\
\quad +  Future&so basically you \tenc{will be satisfied} .\\
\midrule[1pt]
Source&it is n't terrible , but it is n't very good either .\\
Human&it is n't perfect , but it is very good .\\
\midrule
FUDGE&its \tenc{good}, but it isn't very good either.\\
\quad + Past&\senc{whether on vacation or in the car,} this hotel isn't terrible, but it isn't\\
\quad + Present&\senc{whether good} the food \tenc{is}n't terrible, but it \senc{isn't very good} either. \tenc{good good}\\
\quad + Future&several locations \senc{aren't} terrible, but it \senc{is} \tenc{good very good good great}!\\
\midrule
Ours&it is n't terrible , but it is \tenc{very good} also .\\
\quad +  Past&it \tenc{was} n't terrible , but it \tenc{was very good and quick} !\\
\quad +  Present&it \tenc{is} n't terrible , but it \tenc{is very good} also .\\
\quad +  Future&it is n't terrible , but it \tenc{would definitely be very good} !\\
\midrule[1pt]
Source&anyway , we got our coffee and will not return to this location .\\
Human&we got coffee and we'll think about going back\\
\midrule
FUDGE&exactly \senc{zero stars} for any way, we got our coffee and will \senc{not return} to this location.\\
\quad + Past&once our coffee and \senc{will not return} to this location.\\
\quad + Present&once, we \senc{got} our coffee and \senc{will not return} to this location.\\
\quad + Future&once again, we got our coffee and \tenc{will} \senc{not return} to this location.\\
\midrule
Ours&anyway , we got our coffee and will \tenc{always return} to this location .\\
\quad +  Past&anyway , we \tenc{got} our coffee and \tenc{delivered} to this \tenc{friendly} location .\\
\quad +  Present&anyway , we \tenc{love our coffee} and \tenc{this location has to be found} .\\
\quad +  Future&anyway , we got our coffee and \tenc{will continue to return} to this location .\\
\midrule[1pt]
Source&this place is a terrible place to live !\\
Human&this place is a great place to live !\\
\midrule
FUDGE&\tenc{great place} to live!\\
\quad + Past&\tenc{great food} and \senc{terrible service}! \senc{[No Tense]}\\
\quad + Present&\tenc{great place} to live! \senc{[No Tense]}\\
\quad + Future&\tenc{great place} to live! \senc{[No Tense]}\\
\midrule
Ours&this place is a \tenc{great place} to live !\\
\quad +  Past&this place \tenc{was a great place} to live !\\
\quad +  Present&this place \tenc{is a great place} to live !\\
\quad +  Future&this place \tenc{would have a great place} to live !\\\bottomrule
    \end{tabular}
    \vspace{-5pt}
    \caption{Examples of text editing with compositional attributes (sentiment and tense) on the Yelp review dataset. Human is the human-annotated reference for sentiment transfer. We mark the failed spans \senc{red} and successful spans \tenc{blue}.}
    \label{tab:example_tst_yelp}
\end{table}

\subsubsection{Results of Text Editing with Single Attribute}
\label{app:example_text_edit_single}
We conduct text editing with a single attribute on both the Yelp review dataset and the Amazon comment corpus. Since both Yelp and Amazon provide 1000 human-annotated sentences, we also calculate reference-BLEU (rBL, BLEU score between output and human-annotated sentences). 

The automatic evaluation results are in Table~\ref{tab:tst_single}. 
Given a pretrained latent model, ours only requires training a classifier of 3.7K parameters and achieves competitive results compared with the strong baselines of many more parameters.  
Regarding the success rate, our method is in the premier league compared to the methods trained with full labeled data. In respect of content preservation, DiRR distinctly outperforms others, since DiRR processes 1.5B trainable parameters and is trained on the full labeled data ($\sim$440K training data), so big data and big models lead to better performance. However, although we follow the few-shot setting (400 training data), ours also performs well in preserving content. Compared with strong baselines, our method achieves competitive results at fluency and input-output alignment (CTC). Additionally, the generated texts closely align with the training texts, as indicated by a lower MAUVE score.

We also perform human evaluations on Yelp to further measure the transfer quality. 
Three people with related experience are invited to score the generated sentences (1 for low quality and 4 for high quality). We then average the scores as the final human evaluation results.
As the human evaluation results are shown in Table~\ref{tab:tst_single}, our \textsc{LatentOps} performs the best.
Some generated examples are provided in Table~\ref{tab:example_tst_yelp_single} (Yelp) and Table~\ref{tab:example_tst_amazon} (Amazon) to further demonstrate the superiority of our method.  One observation is that our method could focus more on logicality and adopt words appropriate to the context.

\begin{table}[ht]
    \centering
    \small
    \begin{tabular}{@{}lccccccccc@{}}
        \toprule
        \multirow{2}{*}{Methods}&Accuracy$\uparrow$&\multicolumn{3}{c}{Content$\uparrow$} &Fluency$\downarrow$&\multirow{2}{*}{MAUVE$\uparrow$} & \multirow{2}{*}{\#Params}& \multirow{2}{*}{\#Data}\\\cmidrule(r){2-2}\cmidrule(r){3-5}\cmidrule(r){6-6}
        & Sentiment& iBL& rBL& CTC&  PPL& & \\ \midrule
        Source& 0.27& 100& 31.4& 0.500&15.9& 0.873&-&-
        \\
        Human& 0.82& 31.9& 100& 0.463&24.5& 0.055&-&-
        \\ \midrule
        B-GST& 0.81& 31.8& 16.3& 0.473&39.5&  0.513&111M & \multirow{5}{*}{Full-data}
        \\
        STrans  & 0.91& 53.2& \underline{24.5} & 0.469&41.0&  0.904&17M&
        \\
        DiRR& \textbf{0.96}&\textbf{61.5}& \textbf{29.8}& \textbf{0.480}&\underline{23.9}&0.809&1.5B &
        \\
        T\&G & 0.88 & 47.6 &21.8 & 0.466& 24.3 &0.934 &63M&
        \\
        FGST & 0.90 & 13.2 &7.6 & 0.450& \textbf{9.3}&0.593 & 26M&
        \\ \midrule
        FUDGE& 0.40 & \underline{57.0} & 18.0 & 0.456 & 39.3&0.011&16.4M &\multirow{2}{*}{\textbf{Few-shot}}
        \\
        Ours& \underline{0.95} & {54.0} &  24.3& \underline{0.474}& 25.9 &0.902&\textbf{3.7K}&
        \\\midrule\midrule
        Source& 0.14& 100& 49.4& 0.425& 26.4& 0.580  &-&-
        \\
        Human& 0.52& 49.7& 100&0.422& 47.2&  0.541 &-&-
        \\ \midrule
        B-GST& 0.62& 52.3& 28.5& \underline{0.425}& \underline{27.7}& 0.528& 111M &\multirow{4}{*}{Full-data}
        \\
        DiRR& 0.60&\underline{68.7}& \textbf{38.2}& 0.424&32.5 &0.536 &1.5B&
        \\
        T\&G & 0.65 & {68.6} &\underline{35.4} & 0.423& 40.9& 0.535 &63M&
        \\
        FGST & \textbf{0.83} & 21.9 &14.0 & \textbf{0.427}& \textbf{13.6}& 0.524 &26M&
        \\ \midrule
        FUDGE& 0.20&\textbf{70.5}&35.1&0.415&49.5&0.544 &16.4M&\multirow{2}{*}{\textbf{Few-shot}}
        \\
        Ours& \underline{0.72} & 53.3 & 28.1& {0.423}& 44.1&0.547  &\textbf{3.7K}&
        \\\cmidrule[\heavyrulewidth]{1-9}
    \end{tabular}
    \begin{tabular}{ccccccc}
     \cmidrule[\heavyrulewidth]{1-7}
     B-GST & STrans & DiRR & T\&G & FGST & FUDGE & Ours  \\\midrule
     2.03 & 2.20 & 3.13 & 2.20 & 1.60 & 1.20 & \textbf{3.27}\\\bottomrule
    \end{tabular}
    \caption{Automatic evaluations of text editing with single attribute on Yelp (top) and Amazon (middle) dataset.
    We mark the number of trainable parameters as \#Params and the scale of labeled data in training as \#Data.  Human evaluation (bottom) statistics on Yelp.}
    \label{tab:tst_single}
\end{table}
\begin{table}[ht]
    \centering
    \footnotesize
    \vspace{-10pt}
    \begin{tabular}{lp{13cm}}
\toprule
Source&so basically tasted watered down .\\
Human&it didn't taste watered down at all.\\
\midrule
B-GST&so basically tasted delicious .\\
STrans&so basically really clean and comfortable .\\
DiRR&so basically tastes delicious .\\
T\&G&everything tasted fresh and tasted delicious .\\
FGST&everything tasted fresh and tasted like watered down .\\
\midrule
FUDGE&once every couple months, we get a new car - so basically tasted watered down.\\
Ours&so basically tasted delicious .\\
\midrule[1pt]
Source&it is n't terrible , but it is n't very good either .\\
Human&it is n't perfect , but it is very good .\\
\midrule
B-GST&best indian food in whole of pittsburgh .\\
STrans&it is n't great , but it is very good atmosphere .\\
DiRR&it is great , but it is very good either .\\
T\&G&it is n't great , but it is n't very good .\\
FGST&the food is n't very good , but it is n't great either .\\
\midrule
FUDGE&its good, but it isn't very good either.\\
Ours&it is n't terrible , but it is very good also .\\
\midrule[1pt]
Source&anyway , we got our coffee and will not return to this location .\\
Human&we got coffee and we'll think about going back\\
\midrule
B-GST&"got our tickets\\
STrans&anyway , we got our coffee and will definitely return to this location .\\
DiRR&anyway , we got our coffee and will definitely return to this location .\\
T\&G&anyway , we got our coffee and we will definitely return in town .\\
FGST&we will return to this location again , and the coffee was great .\\
\midrule
FUDGE&exactly zero stars for any way, we got our coffee and will not return to this location.\\
Ours&anyway , we got our coffee and will always return to this location .\\
\midrule[1pt]
Source&this place is a terrible place to live !\\
Human&this place is a great place to live !\\
\midrule
B-GST&this place is my new favorite place in phoenix !\\
STrans&this place is a great place to live !\\
DiRR&this place is a great place to live !\\
T\&G&this place is a great place to go !\\
FGST&this place is a great place to live .\\
\midrule
FUDGE&great place to live!\\
Ours&this place is a great place to live !\\
\midrule[1pt]
Source&they are so fresh and yummy .\\
Human&they are not fresh or good .\\
\midrule
B-GST&we are so lazy they need .\\
STrans&they are so dry and sad .\\
DiRR&they are not so fresh and yummy .\\
T\&G&they are not yummy .\\
FGST&it 's so bland and they are tiny .\\
\midrule
FUDGE& mushy rice with egg rolls and a side of egg rolls.\\
Ours&they are just a few and too sour .\\
\midrule[1pt]

Source&i highly recommend this salon and the wonderfully talented stylist , angel .\\
Human&i don't recommend this salon because the artist had no talent.\\
\midrule
B-GST&"i was disappointed to write the salon and the stylist \\
STrans&i was hate this salon and the sloppy dead dead example , angel .\\
DiRR&i would not recommend this salon and the wonderfully incompetent stylist , angel .\\
T\&G&i hate this salon and not wonderfully talented stylist , angel .\\
FGST&i would not recommend this salon to anyone who hates hair , and eyebrow .\\
\midrule
FUDGE&in't a big fan of chain places, but i highly recommend this salon and the wonderfully talented\\
Ours&i would never recommend this salon and the most pathetic stylist named cynthia .\\
\bottomrule
    \end{tabular}
    \vspace{-5pt}
    \caption{Examples of text editing with single attribute on Yelp review dataset. 
    }
    \label{tab:example_tst_yelp_single}
\end{table}

\begin{table}[ht]
    \centering
    \footnotesize
    \vspace{-10pt}
    \begin{tabular}{ll}
\toprule
Source&this is honestly the only case i ve thrown away in the garbage .\\
Human&this is honestly the only case i've kept for so long.\\
\midrule
B-GST&this is honestly the only case i ve put away in the dishwasher .\\
DiRR&this is honestly the only case i ve thrown away in the fridge .\\
T\&G&if your knives had a kickstand on the plate it won t lock down .\\
FGST&it won t slide down on the counter if you have a holder .\\\midrule
FUDGE&this is honestly the only case i ve thrown away in the garbage.\\
Ours&this is honestly the only case i ve saved in the kitchen .\\
\midrule[1pt]
Source&there was almost nothing i liked about this product .\\
Human&there was few features i liked about this product\\
\midrule
B-GST&there was almost no dust i liked about this .\\
DiRR&it was almost perfect for my needs .\\
T\&G&and , there were no where we liked about this pan .\\
FGST&we ve had this for many years , and there are many things about it .\\\midrule
FUDGE&there was almost nothing i liked about be be be and this product.\\
Ours&there is almost all i liked this nice product .\\
\midrule[1pt]
Source&this is not worth the money and the brand name is misleading .\\
Human&this is worth the money and the brand name is awesome.\\
\midrule
B-GST&this is worth the money and the brand name is great .\\
DiRR&this is the perfect size and the price is right .\\
T\&G&i won t be buying any more in the dishwasher .\\
FGST&i won t be buying any more in the future .\\\midrule
FUDGE&this is not worth the money and and be misleading.\\
Ours&this is worth the money and the brand is awesome as the apple .\\
\midrule[1pt]
Source&i ve used it twice and it has stopped working .\\
Human&used it without problems\\
\midrule
B-GST&i ve used it twice and it has held up .\\
DiRR&i ve used it twice and it has worked .\\
T\&G&i ordered num\_num and find this to be a great little mistake .\\
FGST&i find this to be a perfect size .\\\midrule
FUDGE&i ve used be great and it has stopped working.\\
Ours&i ve used it twice and it has still working .\\
\midrule[1pt]
Source&but this one does the job very nicely .\\
Human&but this one does the job well enough\\
\midrule
B-GST&but this one fit the very nicely .\\
DiRR&but this one does the job very poorly .\\
T\&G&plus its from amazon and amazon wouldn t put their name on this game .\\
FGST&shame on amazon and wouldn t buy from amazon .\\\midrule
FUDGE&but this one does the job very nicely.\\
Ours&but this one does the job very negatively .\\
\midrule[1pt]
Source&as stated by the many reviews ,  this is an exceptinal carpet cleaner .\\
Human&as stated by the many reviews , this is a discreet carpet cleaner\\
\midrule
B-GST&as stated by the many reviews , this is an excellent game .\\
DiRR&as stated by the many reviews , this is an exceptinal .\\
T\&G&i also love it because the jar is useless .\\
FGST&i also love the scent because it is plastic .\\\midrule
FUDGE&as stated by the many reviews there will not disappoint there will not disappoint \\
Ours&as stated by the many reviews this is an exceptional poor carpet .\\
\midrule[1pt]
Source&unless you have very small or very large hands it is comfortable to use .\\
Human&unless you have normal sized hands it is uncomfortable to use.\\
\midrule
B-GST&unless you have very small hands or very large hands it is useless .\\
DiRR&unless you have very small or very large hands it is uncomfortable to use .\\
T\&G&not worth these alot and they taste great .\\
FGST&they work alot better than these patches .\\\midrule
FUDGE&unless you have very small or very largest paws there will not a problem.\\
Ours&unless you have very small or very large hands it might be worse .\\
\bottomrule
    \end{tabular}
    \caption{Examples of text editing with single attribute on Amazon comment corpus.}
    \label{tab:example_tst_amazon}
\end{table}

\subsection{Ablation Study: Comparison with SGLD and SDE}
\label{app:compare_sde}
In order to show the superiority of the ODE sampler introduced in \S\ref{sec:ode_sampler}, we compare with Stochastic Gradient Langevin Dynamics (SGLD) and Predictor-Corrector sampler with VP-SDE. The automatic evaluation results are shown in Table~\ref{tab:camparison_sde}. The ODE sampler has the best trade-off between diversity and fluency based on the premise of the success rate. 
\begin{table*}[ht]
\small
\centering
\begin{tabular}{llcccccc}
\toprule
Attributes           & Samplers    & Sentiment$\uparrow$ & Tense$\uparrow$ & Formality$\uparrow$ & G-Mean$\uparrow$& Fluency (PPL)$\downarrow$ &Diversity (sBL)$\downarrow$ \\
\midrule
\multirow{3}{*}{Sentiment}
                 & SGLD &0.64 &-&-&0.64 &\textbf{2.0} & 96.6\\
                 & SDE & \underline{0.82} & -&-& \underline{0.82}& 63.8 & \textbf{6.3} \\
                 & ODE &\textbf{0.99} &-&-&\textbf{0.99} & \underline{30.4} & \underline{13.0} \\
                 \cmidrule{2-8}
\multirow{3}{*}{\quad + Tense} 
                 & SGLD &0.61 &\underline{0.68}&-& 0.644&\textbf{1.9} & 97.8\\
                 & SDE & \underline{0.79} & 0.61&-& \underline{0.692}& 60.6 & \textbf{6.8} \\
                 & ODE &\textbf{0.98}& \textbf{0.93}& -&\textbf{0.951}& \underline{25.2}& \underline{19.7}\\
                 \cmidrule{2-8}
\multirow{3}{*}{\quad +Formality} 
                 & SGLD &0.52 &0.44&\underline{0.82}& 0.573&\textbf{2.3} & 96.8\\
                 & SDE & \underline{0.77} & \underline{0.60}&0.67& \underline{0.675}& 62.5 & \textbf{6.7} \\
                 & ODE &\textbf{0.97}& \textbf{0.92}& \textbf{0.93}&\textbf{0.937}   & \underline{25.8}& \underline{21.1}\\
                 \midrule
\end{tabular}
\caption{Comparison of different sampling method.}
\label{tab:camparison_sde}
\end{table*}

SGLD could generate high quality sentences, but all the sentences contain the similar content, for example: "\texttt{awesome food is great as always !}", "\texttt{great food is awesome as always !}", "\texttt{great food is awesome and always good !}", "\texttt{great place for your haircut .}" and "\texttt{great place with typically no bacon .}". Therefore, it performs the worst in the perspective of diversity. Also, the success rate is at a low level because of the sensitivity and instability of LD (\S\ref{sec:bg_ebms}).

Contrary to SGLD, the SDE sampler cannot guarantee the fluency of the generated sentences, although diversity is good.

We also compute the generation time of different sampling methods as shown in Table~\ref{tab:time_comsumed_sampler}. Combining the automatic evaluation results, sampling by ODE sampler gives the best trade-off among various aspects.

\begin{table}[ht]
    \centering
    \begin{tabular}{cccc}
    \toprule
    Samplers & SGLD & SDE & Ours \\\midrule
    Time & 5.1s (0.93x) & 15.6s (2.85x) & 5.5s (1x)\\\bottomrule
    \end{tabular}
    \caption{Results of generation time of different samplers.}
    \label{tab:time_comsumed_sampler}
\end{table}

\begin{table}
\small
 \setlength\tabcolsep{3pt}
    \centering
    \begin{tabular}{p{0.05\linewidth}p{0.85\linewidth}}
    \toprule
    Initial & Keywords\\
    \midrule
    a & accommodate add afternoon agree airport ambiance ambience amount animal answer anyone anything apartment apologize apology appetizer appointment area arizona arrive art ask atmosphere attention attitude auto average avoid az \\
    \midrule
    b & baby back bacon bag bagel bakery bar bartender base bathroom bbq bean beat become bed beef beer begin believe bell bike bill birthday biscuit bit bite book bottle bowl box boy boyfriend bread breakfast bring brunch buck buffet building bun burger burrito business butter buy \\
    \midrule
    c & cab cafe cake call car card care carry case cash cashier center chain chair chance change charge charlotte check cheese chef chicken child chili chip chocolate choice choose city class cleaning close club cocktail coffee color combo come company condition consider contact continue cook cooky corn cost counter couple coupon course cover crab crave cream credit crew crispy crowd crust cup curry customer cut \\
    \midrule
    d & date daughter day deal dealership decide decor deli deliver delivery dentist department deserve desk dessert detail diner dining dinner dip discount dish do doctor dog dollar donut door downtown dress dressing drink drive driver drop \\
    \midrule
    e & eat egg employee enchilada end entree environment establishment evening event everyone everything expect expectation experience explain eye \\
    \midrule
    f & face facility fact family fan fee feel feeling felt fill find finish fish fit fix flavor flight floor flower folk follow food foot forget friday friend front fruit fry furniture future \\
    \midrule
    g & game garden get gift girl give glass go god grab greet grill grocery ground group guess guest guy gym gyro \\
    \midrule
    h & hair haircut half hand handle happen have head hear heart help hit hold hole home homemade honey hope hospital hostess hotel hour house husband \\
    \midrule
    i & ice idea include ingredient inside item \\
    \midrule
    j & job joint juicy \\
    \midrule
    k & keep kid kind kitchen know \\
    \midrule
    l & lady leave let lettuce level life light line list listen live lobster location look lot lunch \\
    \midrule
    m & mac machine madison make mall man management manager manicure manner margarita mark market massage matter meal mean meat meatball medium meet melt member mention menu mile min mind mine minute mix mom money month morning mouth move movie mushroom music \\
    \midrule
    n & nail name need neighborhood night none noodle notch nothing notice number nurse \\
    \midrule
    o & occasion offer office oil ok okay omelet one onion online open opinion option orange order organize others overcook overprice own owner \\
    \midrule
    p & pack pad pancake park parking part party pass pasta patio pay pedicure people pepper person pet phoenix phone pick picture pie piece pittsburgh pizza place plan plate play please plenty point pool pork portion potato practice prepare price pricing process produce product provide purchase put \\
    \midrule
    q & quality question quick quote \\
    \midrule
    r & ranch rate rating read reason receive refill relax remember rent repair replace request reservation resort rest restaurant result return review rib rice ride ring rock roll room run rush \\
    \midrule
    s & salad sale salmon salon salsa salt salty sandwich saturday sauce sausage save saw say schedule school scottsdale seafood season seat seating section see seem selection sell send sense serve server service set share shoe shop shopping shot show shrimp side sign sit size slice soda someone something son sound soup space speak special spend spice spicy spinach sport spot spring staff stand standard star starbucks start state station stay steak step stick stock stop store story street strip stuff style stylist sub suggest summer sunday suppose surprise sushi \\
    \midrule
    t & table taco take talk taste tasty tea team tech tell thai thanks theater thing think throw time tip tire toast today tomato ton tonight topping tortilla touch town treat trip try tuna turn tv type \\
    \midrule
    u & understand update use \\
    \midrule
    v & valley value vega vegetable veggie vehicle venue vet vibe view visit \\
    \midrule
    w & waffle wait waiter waitress walk wall want wash watch water way wedding week weekend while wife window wine wing wish woman word worker world wrap write \\
    \midrule
    y & year yelp yesterday yummy \\
   \bottomrule
    \end{tabular}
    \caption{All keywords. Sort in alphabetical order.}
    \label{tab:all_keywords}
\end{table}

\end{document}